\documentclass{article}

\usepackage[final,nonatbib]{neurips_2022_ml4ad}

\usepackage[utf8]{inputenc} %
\usepackage[T1]{fontenc}    %
\usepackage{hyperref}       %
\usepackage{url}            %
\usepackage{booktabs}       %
\usepackage{amsfonts}       %
\usepackage{nicefrac}       %
\usepackage{microtype}      %

\usepackage{graphicx}
\usepackage{amsmath}
\usepackage{amssymb}
\usepackage{microtype}
\usepackage{soul}
\usepackage{xspace}
\usepackage[usenames,dvipsnames]{xcolor}
\usepackage{pgfplots}
\usepackage{pgfplotstable}
\pgfplotsset{compat=1.15}
\usepackage{multirow}
\usepackage{subfigure}
\usepackage{enumitem}
\usepackage[numbers]{natbib}

\usepackage{capt-of}

\newcolumntype{C}[1]{>{\centering\let\newline\\\arraybackslash\hspace{0pt}}m{#1}}

\newcommand{\ours}{SafePathNet\xspace}
\newcommand{\anchors}{learnable query embeddings\xspace}

\newcommand{\ablationsubsection}[1]{\paragraph{#1}}

\newcommand{\loss}{\mathcal{L}}
\newcommand{\mincost}{MinCost\xspace}
\newcommand{\mincostcc}{MinCostCC\xspace}
\newcommand{\ilk}{events per 1k miles\xspace}
\newcommand{\ablationmarkersize}{0.9mm}
\newcommand{\ablationmarkerlinesize}{0.6mm}
\DeclareMathOperator*{\argmin}{arg\,min}

\renewcommand{\citep}[1]{\cite{#1}}
\renewcommand{\citet}[1]{\cite{#1}}
\newcommand{\ie}{i.e.\xspace}
\newcommand{\shorturl}{https://woven.mobi/safepathnet}

\newcommand{\timearrow}{
  \begin{tikzpicture}
    \draw[thick,->] (0,0) -- (0.75\linewidth,0) node[midway,above] {Time};
  \end{tikzpicture}
}

\definecolor{mygray}{rgb}{0.5,0.5,0.5}

\def\singlecol{1}

\title{Safe Real-World Autonomous Driving by Learning\\to Predict and Plan with a Mixture of Experts}

\author{
  Stefano Pini$^1$,\hspace{2mm} Christian S. Perone$^1$,\hspace{2mm} Aayush Ahuja$^2$,\hspace{2mm} Ana Sofia Rufino Ferreira$^2$,\\
  \textbf{Moritz Niendorf$^2$,\hspace{2mm} Sergey Zagoruyko$^1$} \\ \\
  \textmd{$^1$Woven Planet United Kingdom Limited}\\
  \textmd{$^2$Woven Planet North America, Inc.}\\
  {\small \texttt{\{firstname\}.\{lastname\}@woven-planet.global}}
}

\begin{document}

\maketitle

\begin{abstract}
The goal of autonomous vehicles is to navigate public roads safely and comfortably. To enforce safety, traditional planning approaches rely on handcrafted rules to generate trajectories. 
Machine learning-based systems, on the other hand, scale with data and are able to learn more complex behaviors.
However, they often ignore that agents and self-driving vehicle trajectory distributions can be leveraged to improve safety.
In this paper, we propose modeling a distribution over multiple future trajectories for both the self-driving vehicle and other road agents, using a unified neural network architecture for prediction and planning. 
During inference, we select the planning trajectory that minimizes a cost taking into account safety and the predicted probabilities.
Our approach does not depend on any rule-based planners for trajectory generation or optimization, improves with more training data and is simple to implement. 
We extensively evaluate our method through a realistic simulator and show that the predicted trajectory distribution corresponds to different driving profiles. We also successfully deploy it on a self-driving vehicle on urban public roads, confirming that it drives safely 
without compromising comfort.
The code for training and testing our model on a public prediction dataset and the video of the road test are available at \url{\shorturl}.
\end{abstract}

\section{Introduction}\label{sec:introduction}

In the last decade, Autonomous Driving (AD) has been largely explored by researchers in academia and industry.
Against expectations of many in the field, Self-Driving Vehicles (SDVs) are still constrained to limited operational domains and not yet ready to be deployed at scale.
Among the AD stack, the weak link appears to be the planning module,
which is where most decision-making takes place.
This component not only has to reason about other road actors'
actions and cover different driving behaviors, it must also guarantee safety and robustness against the long tail distribution of driving scenarios.

Traditional rule-based systems~\citep{paden2016survey} addressed the planning task by defining progressively larger sets of hand-crafted rules,
but they proved to scale poorly to complex or unfamiliar driving scenarios. 
While several data-driven approaches~\citep{wulfmeier2017large,bansal2018planning-5,zeng2019end,sadat2020perceive,cui2021lookout} were presented in recent years to improve scalability
and performance by leveraging large datasets,
they either approach planning as a unimodal trajectory forecasting problem, %
lack safety checks,
or are computationally inefficient.
Recently, \citet{Vitelli2021SafetyNetSP} presented a hybrid two-stage approach
consisting of a ML-based planner for trajectory forecasting and a safety fallback layer.
However, the method relies on both an external prediction module and a rule-based trajectory generator that does not learn nor improve with data.

In this paper, we propose \ours, a 
ML-based
prediction and planning system that leverages data uncertainty 
and scales with training data, while 
ensuring
a safety profile similar to previous methods. %
\ours jointly models prediction and planning as distributions of future trajectories, and leverages predictions to improve safety at inference time.
Exploiting the powerful attention mechanism of the Transfomer module~\cite{NIPS2017_3f5ee243}, we propose a deep neural network architecture that predicts diverse SDV's trajectories and other agents' future locations given a vectorized representation of the scene.
To model data uncertainty, we use a Mixture of Experts (MoE) approach~\citep{jacobs1991adaptive} that learns a distribution over trajectories.
To improve safety of the SDV's plan, we perform cost-sensitive selection from the trajectory distribution according to the road agents' predicted locations to avoid collisions.
This approach to safety is efficiently performed within the model itself, scales with data, and does not rely on any additional external signals, as \ours selects the trajectory from the predicted distribution.
Simulation and on-road tests show that the proposed approach models and exploits different driving profiles to improve planning safety without substantially impacting comfort.

In summary, our contributions are:
\ifx\singlecol\undefined\begin{itemize}
\else\begin{itemize}[leftmargin=2em,topsep=0em,itemsep=0pt]\fi
  \item
    We propose to model the distribution of future
trajectories of agents and the SDV using a mixture of experts in a unified neural network for prediction and planning;

  \item
  We present an efficient and easy to implement decision-making approach that leverages the predicted trajectories and associated probabilities to improve safety by reducing collisions between the SDV and other road agents;
  \item
  We extensively validate our proposal in a realistic closed-loop simulator and deploy it on an SDV driving on public roads, confirming its effectiveness and safety;
  \item
  The code and hyperparameters for training and testing our model on a public prediction dataset will be made publicly available.
\ifx\singlecol\undefined\end{itemize}\else\end{itemize}\fi

\begin{figure}[t]
    \centering
    \includegraphics[width=\textwidth, trim=0 20 0 0]{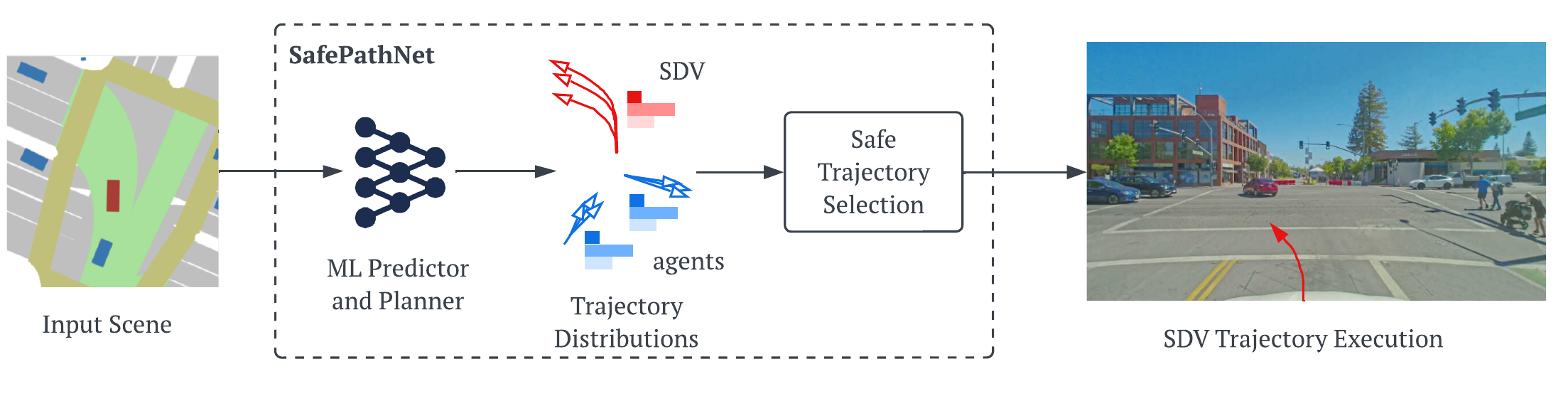}
    \caption{High-level overview of \ours, a ML approach improving on-road safety of self-driving vehicles (SDVs).
    A neural network receives a vectorized scene representation, combining map, road agent  and SDV states.
    It then predicts a set of future SDV driving plans (in red) and a set of agent future trajectories (in blue) with associated probabilities.
    A safe driving trajectory is selected given this data, and executed by the SDV in the real world.}
    \label{fig:high_level_arch}
\end{figure}

To the best of our knowledge, this is the first paper that combines a MoE approach with a decision-making strategy that 
uses the predicted distribution to drive safely on public roads  (Figure \ref{fig:high_level_arch}).

\section{Related work}\label{sec:related}
Although many data-driven systems
have been 
\ifx\singlecol\undefined 
developed in the past years~\citep{bansal2018planning-5,Gao_2020_CVPR}
\else
recently developed~\citep{bansal2018planning-5,Gao_2020_CVPR}
\fi
exploiting breakthroughs in Deep Learning methods, many challenges remain open for the deployment of these systems in the real world.
Deep learned systems can scale and improve with data and are increasingly favored over hand-engineered approaches that do not scale in engineering effort. However, they also present new challenges, e.g.~how to embed causal relationships and behaviors such as collision avoidance.

Data-driven approaches to planning can be grouped according to their training paradigm, such as Imitation Learning (IL)~\citep{zheng2021imitation} and Reinforcement Learning (RL)~\citep{Sutton1998}.
Among IL methods, Behavioral Cloning~\citep{bain1995framework} is one of the most used approaches of imitation, dating back to ALVINN~\citep{Pomerleau1989} and showing recent developments such as in~\citep{bansal2018planning-5,zeng2019end,Vitelli2021SafetyNetSP,Hawke2020UrbanDW,bojarski2016end2end}.
Although imitation approaches showed significant progress, the covariate shift induced by the policy is still an open problem that can make the model perform poorly during deployment.  %
On the other hand, Reinforcement Learning~\citep{Sutton1998} has reached many milestones on a large set of simulated environments. However, its uptake in real-world problems has been slower than expected~\citep{challenges2021RL}. While several methods have been proposed in recent years~\citep{shalev2016safe,riedmiller2007learning,kendall2019learning}, they only make use of simulated environments
or show very limited and constrained private-road testing, limiting their application in the real world.
PredictionNet~\citep{kamenev2022predictionnet} show on-road tests too, but the RL policy only controls the SDV longitudinal speed.
When tested in the real world, most of the aforementioned approaches make mistakes, mainly due to covariate shift.
For this reason, safety guarantees are paramount for deployment in real scenarios.
The most similar approach to our method is SafetyNet~\citep{Vitelli2021SafetyNetSP},
where a two-stage pipeline is used to perform rule-based safety checks over a ML-based trajectory.
SafetyNet depends on an external rule-based planner based on~\citet{werling2012optimal} and does not scale with more data as the trajectory generator is 
not learned.
On the contrary, our approach learns to predict diverse trajectories from data, and exploits the SDV and road agent predictions to improve the safety of the planner.

Other works~\citep{cui2018multimodal,minh2019multimodal,casas2020implicit,cui2021lookout,Varadarajan2021MultiPathEI,kamenev2022predictionnet,amini2019variational} propose the 
generation of multiple trajectories
in the context of the SDV motion planning, localization, or road agent prediction.
Some of these works~\citep{cui2018multimodal, minh2019multimodal} produce predictions independently of each other while others~\citep{casas2020implicit,cui2021lookout} produce scene-level consistent predictions.
While the former approaches
can generate unsafe SDV trajectories colliding with other road agents,
the latter 
often suffer from sample inefficiency due to the large set of possible futures.
In comparison, our approach simultaneously predicts a diverse set of SDV and road agent trajectories and associated probabilities and leverages them to increase the safety of the planner during inference in real-time.

\section{Methodology}\label{sec:method}
In this section, we first present the multimodal architecture and the training procedure we use to jointly predict future trajectories of other road agents and plans of the SDV.
Then, we present a simple-yet-effective approach to improve  safety of the planner during inference. We propose a cost-sensitive selection over the predicted SDV trajectory distribution exploiting road agents future locations to avoid collisions with them.

\subsection{Model architecture}

\begin{figure*}
    \centering
    \includegraphics[width=\linewidth, trim=0 15 0 0]{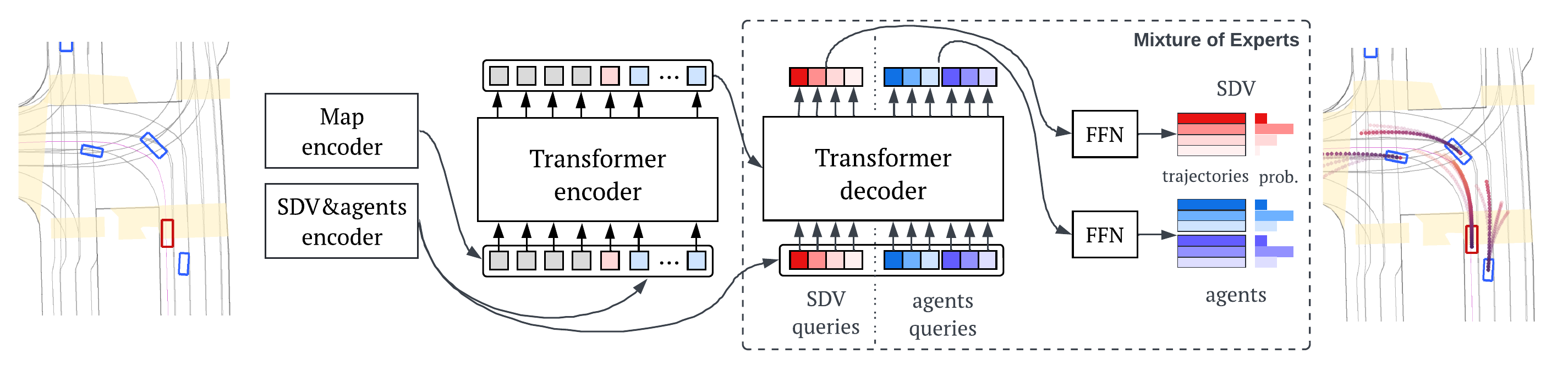}
    \caption{\ours predicts SDV and road agents future
    trajectories given a vectorial representation of input scene.
    Firstly, map, SDV and agent features are encoded independently by PointNet-like networks.
    Then, their outputs are processed by a Transformer encoder-decoder network.
    Finally, decoder outputs are processed by FFNs predicting SDV and agent future trajectories together with probability distributions.
    Additionally, we use a kinematic model after SDV FFN, omitted for clarity.
    }
    \label{fig:detail_arch}
    \ifx\singlecol\undefined\vspace{-10pt}\fi
\end{figure*}

We address the joint tasks of prediction and planning with the neural architecture shown in Figure~\ref{fig:detail_arch},
inspired by previous approaches~\cite{Gao_2020_CVPR,Vitelli2021SafetyNetSP}.
Differently from such prior work, which mostly use unimodal predictions, we model the multimodal trajectory distribution using a Mixture of Experts (MoE) approach.
That is, the deep model predicts different trajectory candidates
and a probability distribution over them.
We can then take advantage of these planning alternatives defining a selection policy to improve driving safety at inference time. %

\textbf{Input and output representation.}
We represent the driving scene in a vectorized format,
and use the following data as input:
    (i) SDV as current pose, speed, acceleration, size, moving/stationary history for $K_s$ seconds;
    (ii) road agents as pose, size, vehicle type 
    for the current frame and the previous $K_a$ seconds;
    (iii) static HD map including lanes, crosswalks, intersections;
    (iv) dynamic map elements including other non-moving obstacles and status of traffic lights;
    (v) goal (route) as the center of the lane that the SDV should follow.
Each input element points are encoded in SDV-centric reference frame and include the element type as additional feature.

The model outputs can be split in planning and prediction output.
The planning output is composed of $N$ SDV future trajectories $\tau^i$
and one probability distribution $p_i = p(\tau^i \,|\, \mathbf{x})$ over them.
Each SDV trajectory $\tau^i$ is defined as a set of $T_s$ discrete states
\begin{equation}
    \tau^i_t = \big\{ x^i_t, y^i_t, \theta^i_t, v^i_t, a^i_t, k^i_t, j^i_t \big\},
\end{equation}
where $t$ represents a timestep in the range $[1, T_s]$, $x,y,\theta$ the pose, $v$ the speed, $a$ the acceleration, $k$ the curvature, and $j$ the jerk.
In practice, the model outputs the jerk and curvature $k^i_t, j^i_t$ for each timestep $t$ and the remaining trajectory features are inferred using a unicycle kinematic vehicle model from the initial state ${x^i_0, y^i_0, \theta^i_0}$.
The probability distribution $p_i = p(\tau^i \,|\, \mathbf{x})$ is defined over the $N$ SDV trajectories $\tau^i$ and can be use to pick the most appropriate one given the current input. We discuss this in detail in Section~\ref{subsec:mode_selection}.

The prediction output is composed of $A \times M$ road agents' future trajectories $\nu_a^j, \, j=1, \dots M$
and $A$ probability distributions $q^a_j = p(\nu_a^j \,|\, \mathbf{x})$ over each set of agent $M$ future trajectories $\nu_a^j$.
Each road agent trajectory $\nu^j$ is instead defined as a set of $T_a$ discrete states
\begin{equation}
    \nu^j_t = \big\{ x^j_t, y^j_t, \theta^j_t \big\},
\end{equation}
where $t$ represents a timestep in the range $[1, T_a]$ and $x,y,\theta$ represent the pose.
Similarly to the planning output, each probability distribution $q^a, \, a = 1, \dots A$ is defined over the $M$ trajectories of the $a$-th agent. The distribution can be used to pick the most appropriate trajectory for each agent.

\ifx\singlecol\undefined
\begin{table*}[t]
  \centering
  \caption{
  Comparison with competitors in closed-loop virtual evaluation. Results are reported in number of events\\per 1k miles with the .95 confidence interval. 
  Lower is better for all metrics.
  }
      \begin{tabular}{l c c c c c}
    \toprule
    \bf{Planner type} & \bf{Estimated Contacts} & \bf{Close calls} & \bf{Discomfort Brakings} & \bf{Passiveness} \\
    \midrule
    ML planner~\citep{Vitelli2021SafetyNetSP}* & 71 \textcolor{mygray}{(52, 97)} & 116 \textcolor{mygray}{ (91, 148)} & 45 \textcolor{mygray}{(31, 67)} & 85 \textcolor{mygray}{(64, 113)} \\
    ML planner + fallback layer~\citep{Vitelli2021SafetyNetSP}* & 42 \textcolor{mygray}{(28, 63)} & 75 \textcolor{mygray}{(55, 101)} & 284 \textcolor{mygray}{(243, 331)} & 737 \textcolor{mygray}{(670, 810)} \\
    \ours (ours) & 44 \textcolor{mygray}{(29, 65)} & 87 \textcolor{mygray}{(66, 116)} & 69 \textcolor{mygray}{(50, 95)} & 133 \textcolor{mygray}{(106, 167)} \\
    \bottomrule
\end{tabular}
  \label{tab:main_safety}
\end{table*}
\fi

\textbf{Architectural details.}
The architecture of \ours{} is similar to those of VectorNet~\cite{Gao_2020_CVPR} and DETR~\cite{detr}, combining an element-wise point encoder~\cite{qi2017pointnet} and a Transformer~\cite{NIPS2017_3f5ee243} (Fig.~\ref{fig:detail_arch}).
The element-wise point encoder consists of two PointNet-like modules that are used to compress each input element from a set of points to a single feature vector of the same size.
A series of Transformer Encoder layers are used to model the relationships between all input elements (SDV, road agents, static and dynamic map, route), encoded by the point encoder.
Then, a series of Transformer Decoders are used to query SDV and agents features.
We make use of a set of learnable embeddings to construct the queries of the Transformer Decoders.
The SDV embeddings from the point encoder are added to the set of $N$ learnable embeddings to obtain a different query for each SDV future trajectory that we aim to predict.
Similarly, $M$ \anchors are used to obtain a variable number of $M$ different queries for each road agent.
This architecture closely resembles DETR object detection network, fitting our set prediction task well.
Here, each query embedding can encode a specific driving behavior corresponding to one expert of the MoE approach.

Finally, an SDV-specific decoder (FFN) converts each SDV feature to a set of control inputs (i.e.~jerk, curvature) and a kinematic decoder
translates them into a future trajectory.
Similarly, an agent-specific decoder converts each agent feature to a future trajectory.
In addition to trajectories, the decoder predicts a logit for each SDV and agent trajectory. For each element, the corresponding logits are converted to a probability distribution over the future trajectories by applying a
softmax function.
All road agents and SDV are modeled independently, but predicted jointly in parallel.

\textbf{Training procedure.}
We adopt imitation learning and define our training objective as minimizing a distance between predicted SDV poses and the ground truth expert trajectories.
Similarly, we minimize distance between predicted and ground truth agents' future trajectories.
We additionally regularize jerk and curvature corresponding to SDV plans.

Our model represents a MoE and predicts multiple trajectories for the SDV and each road agent, corresponding to $N$/$M$ experts, and a probability distribution over each trajectory set, corresponding to expert selection.
To train the experts and expert selection jointly while avoiding mode collapse, we use a winner-takes-all approach, somewhat similar to previous methods~\cite{cui2018multimodal}. 
Similarly to DETR~\cite{detr}, we formulate a matching cost
between predicted and target trajectories and probabilities,
making the expert with minimal cost the winner.
Without loss of generality, in the following we describe the loss applied to one sample of the SDV planning, which is similarly applied to agent prediction too.

In details, we compute matching cost for each trajectory then select one trajectory according to
\ifx\singlecol\undefined
    \begin{multline}
        i^* = \argmin_i \; \loss_\text{IL}^i + \lambda (1- p_i), \quad\; \\
    \text{where} \;\; \loss_\text{IL}^i = \sum_{t=1}^{T} ||\tau_t^i - \hat{\tau_t}||_1 + \beta \loss_\text{reg}^i,
        \label{eq:assignment_cost}
    \end{multline}
\else
    \begin{equation}
        i^* = \argmin_i \; \loss_\text{IL}^i + \lambda (1- p_i), \quad\;
    \text{where} \;\; \loss_\text{IL}^i = \sum_{t=1}^{T} ||\tau_t^i - \hat{\tau_t}||_1 + \beta \loss_\text{reg}^i,
        \label{eq:assignment_cost}
    \end{equation}
\fi
where $p_i$ is the predicted probability for the trajectory $\tau^i$, $\hat{\tau}$ is the ground truth trajectory, $\lambda$ and $\beta$ are weighting factors, and $\loss_\text{reg}$ is a regularization loss.
Then, we minimize the following loss
\begin{equation}
    \loss = \loss_\text{IL}^{i^*} + \mu \loss_{\text{NLL}}^{i^*}, \quad\; \text{where} \;\;  \loss_\text{NLL}^{i^*} = -\log \, p_{i^*},
    \label{eq:main_loss}
\end{equation}
that takes into account the selected trajectory $\tau^{i^*}$ and combines the imitation and the matching loss.

\subsection{Inference-time planning policy}\label{subsec:mode_selection}

At inference time, we leverage the diverse predicted trajectories $\tau_i$ to compute the cost $c_i$ of executing each of them. Then, we select the trajectory $\hat{i} = \argmin_i c_i$ with minimum cost.

\textbf{MinCost policy.}
Given the predicted SDV’s trajectories and the associated probabilities, we can simply define the cost to be negatively proportional to the predicted 
probability: $c_i = - p_i$.
However, this trivial approach 
ignores other road agents' future locations and thus may lead to collisions,
as models can still predict colliding trajectories even when trained with auxiliary collision losses~\citep{bansal2018planning-5}.

\textbf{MinCostCC policy.}
In the work of \citet{Vitelli2021SafetyNetSP}, a safety layer is added to enforce safety constraints, e.g. collision avoidance, and legality constraints, e.g. respecting road rules, by checking the SDV trajectory predicted by the ML-based planner.
In case of violations, a fallback trajectory generated by a rule-based planner is used instead of the ML-based one.
While this approach was shown to substantially increase safety, the fallback trajectories are generated using a rule-based system that 
does not improve
with data. 
As a result, the performance of the ML planner — which can improve by training on more data — is capped by the hand-engineered rule-based planner.

To overcome these limitations, we propose to improve the safety of the planner by leveraging the predicted SDV trajectory distribution and agents' predictions instead of relying on external modules.
To this end, we first perform a Collision Check between each future SDV trajectory $\tau^i$ and the most probable predicted agents' locations $\nu^{j*}$ by means of overlap between their bounding boxes.
We use the Separating Axis Theorem (SAT)~\citep{boyd2004convex} for efficient computation.
Then, we extend the cost defined previously by adding a cost for any potential collision with other agents:
\begin{equation}
    c_i = - p_i - \alpha \overline{t}_i
\end{equation}
where $\alpha$ is a fixed penalty term and $\overline{t}_i$ is the timestep of the first predicted collision.
In other words, we penalize SDV trajectories that are predicted to collide with road agents most probable futures. If the predicted set of trajectories contains at least one collision-free trajectory or trajectories with collisions further ahead in the predicted horizon, then our approach can improve the safety of the planner.

\section{Experimental Evaluation}\label{sec:results}
In this section, we first present the proprietary dataset we used to train and 
validate \ours, and our experimental setting.
Then, we present results of our evaluation, which include simulation, comparison of prediction performance on a public dataset, and public road tests. We also present an ablation study of our major components.
While we are unable to release our proprietary simulator and dataset, we plan to release code and hyperparameters for training and testing our model on a public dataset to facilitate reproducibility.

\subsection{Dataset and Experimental Setting}  %
We use a proprietary dataset collected on an SDV platform in challenging urban areas of San Francisco and Palo Alto to train and evaluate our models.
The dataset is composed of scenes spanning from 10 to 30 seconds and containing SDV recorded trajectory, HD map, and outputs of a proprietary perception system.
Training and validation sets have 270 and 60 hours of driving respectively.

We trained the neural model on our training dataset applying the objectives presented in Section~\ref{sec:method}.
We used the Adam optimizer and a base learning rate of $10^{-3}$ with cosine update schedule~\citep{loshchilov-ICLR17SGDR}.
We apply synthetic perturbations to the training data~\citep{Ross2011ARO,Vitelli2021SafetyNetSP} to improve closed-loop performance.
We trained for $40$ epochs on a cluster with 8 nodes having 8 GPUs each, using a batch size of $64$ samples per replica.
Our model has 3 Transformer encoder and 3 decoder layers, with 5.7M parameters in total. Training takes about 5 hours.
\ifx\singlecol\undefined\else For more information, please refer to the Appendix, Section~\ref{appendix:details}.\fi

\ifx\singlecol\undefined\else
\begin{table}[t]
  \centering
  \caption{
  Comparison with competitors in closed-loop virtual evaluation. Results are reported in number of events per 1k miles with the .95 confidence interval.
  Lower is better for all metrics.
  }
  \resizebox{\linewidth}{!}{
      \begin{tabular}{l c c c c c}
    \toprule
    \bf{Planner type} & \bf{Estimated Contacts} & \bf{Close calls} & \bf{Discomfort Brakings} & \bf{Passiveness} \\
    \midrule
    ML planner~\citep{Vitelli2021SafetyNetSP}* & 71 \textcolor{mygray}{(52, 97)} & 116 \textcolor{mygray}{ (91, 148)} & 45 \textcolor{mygray}{(31, 67)} & 85 \textcolor{mygray}{(64, 113)} \\
    ML planner + fallback layer~\citep{Vitelli2021SafetyNetSP}* & 42 \textcolor{mygray}{(28, 63)} & 75 \textcolor{mygray}{(55, 101)} & 284 \textcolor{mygray}{(243, 331)} & 737 \textcolor{mygray}{(670, 810)} \\
    \ours (ours) & 44 \textcolor{mygray}{(29, 65)} & 87 \textcolor{mygray}{(66, 116)} & 69 \textcolor{mygray}{(50, 95)} & 133 \textcolor{mygray}{(106, 167)} \\
    \bottomrule
\end{tabular}
  }
  \label{tab:main_safety}
\end{table}
\fi

\subsection{Results from Simulation}

We first evaluate our approach in simulation, in two different settings, relying on recorded scenes from the real world.
To compare with other methods, we use a
high-fidelity simulator which simulates the SDV physics through a kinematic 
model and log-replay road agents applying a longitudinal speed control to avoid front/side collisions. The SDV drives for approximately 580 miles.
For ablation studies, we use an efficient simulator where we do not apply any kinematic constraints, even though the planner output is constrained by a kinematic model, and log-replay road agents. In this case, the SDV drives for about 340 miles.
The simulators use a high-level representation of the scene; they do not simulate raw sensor data.
In this Section, we compare with our implementation inspired by SafetyNet~\citep{Vitelli2021SafetyNetSP}, since we do not have access to the original implementation. 

The reported metrics are: (i) Estimated Contacts (ECs), \ie number of times the SDV bounding box is closer to a road agent than 5cm or static obstacle by 1cm; (ii) Close Calls, \ie number of times the SDV time-to-collision is less than 1.5 seconds, or SDV time headway is less than 1 second; (iii) Discomfort Brakes (DBs), \ie number of excessive braking events; (iv) Passiveness, \ie number of times SDV drives slower than in the log by at least 5 m/s while being behind the log.
These metrics are computed as the number of events per 1k driven miles.
We also report minADE and minFDE, \ie minimum average and final distance error.

\begin{figure}
    \centering
    \resizebox{0.95\linewidth}{!}{
    {
      \small
      \ifx\singlecol\undefined
        \input{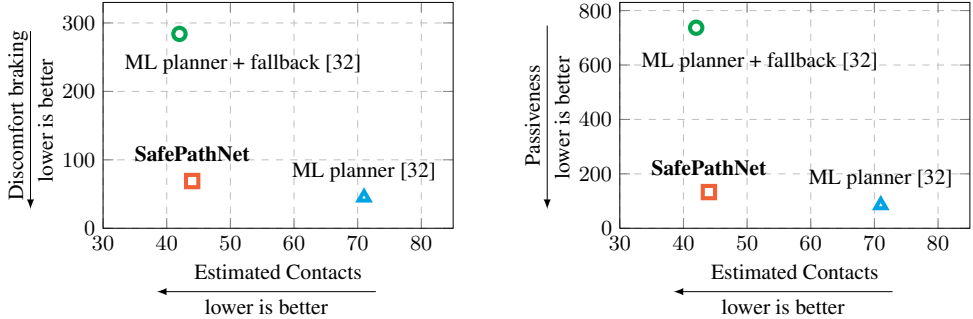}
      \else
        \begin{tikzpicture}
  \begin{axis}[
    name=SimDBAxis,
    xmin=30, xmax=85,
    ymin=0, ymax=330,
    xlabel={Estimated Contacts},
    ylabel={Discomfort braking \\ },
    xlabel style={align=center},
    ylabel style={align=center},
    ymajorgrids=true,
    xmajorgrids=true,
    grid style=dashed,
    width=0.48\linewidth,
    height=0.35\linewidth,
  ]

    \node[label={\bf \ours}] at (44, 75) {};
    \addplot[
        color=RedOrange,
        line width=\ablationmarkerlinesize,
        mark=square,
        mark size=\ablationmarkersize,
        ]
        coordinates {(44, 69)};

    \node[label={ML planner~\cite{Vitelli2021SafetyNetSP}}] at (71, 45) {};
    \addplot[
        color=Cerulean,
        line width=\ablationmarkerlinesize,
        mark=triangle,
        mark size=\ablationmarkersize,
        ]
        coordinates {(71, 45)};

    \node[label={}] at (52, 240)
        {ML planner + fallback~\cite{Vitelli2021SafetyNetSP}};
    \addplot[
        color=Green,
        line width=\ablationmarkerlinesize,
        mark=o,
        mark size=\ablationmarkersize,
        ]
        coordinates {(42, 284)};
        
  \end{axis}

  \draw [-latex] ([yshift=-0.5ex, xshift=-8ex]SimDBAxis.outer south east)
    --node[below]{lower is better} ([yshift=-0.5ex, xshift=16ex]SimDBAxis.outer south west);
  \draw [-latex] ([yshift=-3ex]SimDBAxis.outer north west) ++(3ex,0) coordinate(start)
    --node[sloped,below, rotate=180] {lower is better} ([yshift=8ex]start |- SimDBAxis.outer south west);

\end{tikzpicture}
        \hspace{0.05\linewidth}
        \begin{tikzpicture}
  \begin{axis}[
    name=SimPassivenessAxis,
    xmin=30, xmax=85,
    ymin=0, ymax=830,
    xlabel=Estimated Contacts,
    ylabel={Passiveness \\ },
    xlabel style={align=center},
    ylabel style={align=center},
    ymajorgrids=true,
    xmajorgrids=true,
    grid style=dashed,
    width=0.48\linewidth,
    height=0.35\linewidth,
  ]

    \node[label={\bf \ours}] at (44, 140) {};
    \addplot[
        color=RedOrange,
        line width=\ablationmarkerlinesize,
        mark=square,
        mark size=\ablationmarkersize,
        ]
        coordinates {(44, 133)};

    \node[label={ML planner~\cite{Vitelli2021SafetyNetSP}}] at (71, 90) {};
    \addplot[
        color=Cerulean,
        line width=\ablationmarkerlinesize,
        mark=triangle,
        mark size=\ablationmarkersize,
        ]
        coordinates {(71, 85)};

    \node[label={}] at (52, 610)
        {ML planner + fallback~\cite{Vitelli2021SafetyNetSP}};
    \addplot[
        color=Green,
        line width=\ablationmarkerlinesize,
        mark=o,
        mark size=\ablationmarkersize,
        ]
        coordinates {(42, 737)};
        
  \end{axis}
  \draw [-latex] ([yshift=-0.5ex, xshift=-8ex]SimPassivenessAxis.outer south east)
    --node[below]{lower is better} ([yshift=-0.5ex, xshift=16ex]SimPassivenessAxis.outer south west);
  \draw [-latex] ([yshift=-3ex]SimPassivenessAxis.outer north west) ++(3ex,0) coordinate(start)
    --node[sloped,below, rotate=180] {lower is better} ([yshift=8ex]start |- SimPassivenessAxis.outer south west);

\end{tikzpicture}
      \fi
    }
    }
    \caption{Comparison of Estimated Contacts against Discomfort Brakes/Passiveness for different planners, in \ilk. Lower is better.
    Our model presents a better trade-off between comfort and safety.
    }
    \label{fig:main_table}
    \ifx\singlecol\undefined\vspace{-10pt}\fi
\end{figure}

We first present results obtained with the high-fidelity simulation.
As can be seen in Table~\ref{tab:main_safety},
the naive ML-based planner experiences a high number of Estimated Contacts and Close calls to other road agents.
We also report Discomfort Brakes (DBs) and Passiveness,
showing the number of excessive braking events and the number of times SDV drives slower than in the log respectively, per 1k miles.
Enabling the rule-based fallback layer on top of ML-based predictions~\cite{Vitelli2021SafetyNetSP}, the safety of the planner is significantly improved, at the expense of passiveness and comfort (see large increase in DBs).
On the contrary, our fully ML-based model presents comparable performance in terms of estimated contacts and close calls 
while having a limited impact to DBs and passiveness.
This result confirms that our approach, leveraging the modelled SDV trajectory distribution and the agent predictions, improves the safety of the planner 
without relying on any external rule-based trajectory generator or impacting the riding comfort significantly (Fig. \ref{fig:main_table}). 

Qualitative samples comparing the two polices in our closed-loop simulator are shown in
Figure~\ref{fig:qualitative}. As highlighted by the green arrows, \mincostcc can successfully avoid collisions with other road agents.
Figure~\ref{fig:qualitative_additional} shows additional qualitative samples obtained using the \mincostcc policy. %
We find that SDV trajectories mostly differ in speed and acceleration profile due to route conditioning, but show diverse curvature when turning and nudging parked vehicles. Agent trajectories vary in both curvature, speed and acceleration profile.

\ifx\singlecol\undefined
\begin{figure}
    \centering
    \footnotesize
    \ifx\singlecol\undefined
    \timearrow
    \begin{tabular}{cccc}
        \vspace{-5pt}\\
        \rotatebox[origin=l]{90}{\hspace{4mm}\mincost} &
        \includegraphics[width=0.26\linewidth]{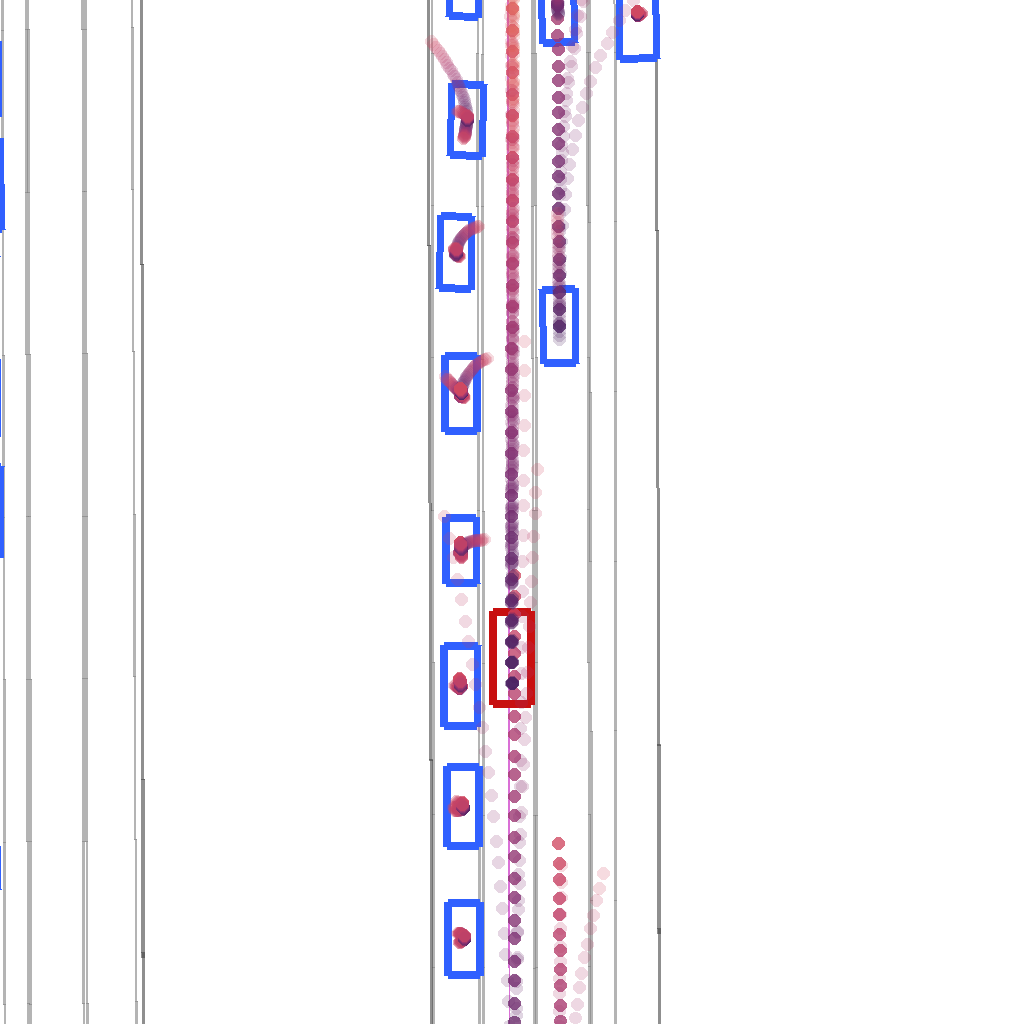} &
        \includegraphics[width=0.26\linewidth]{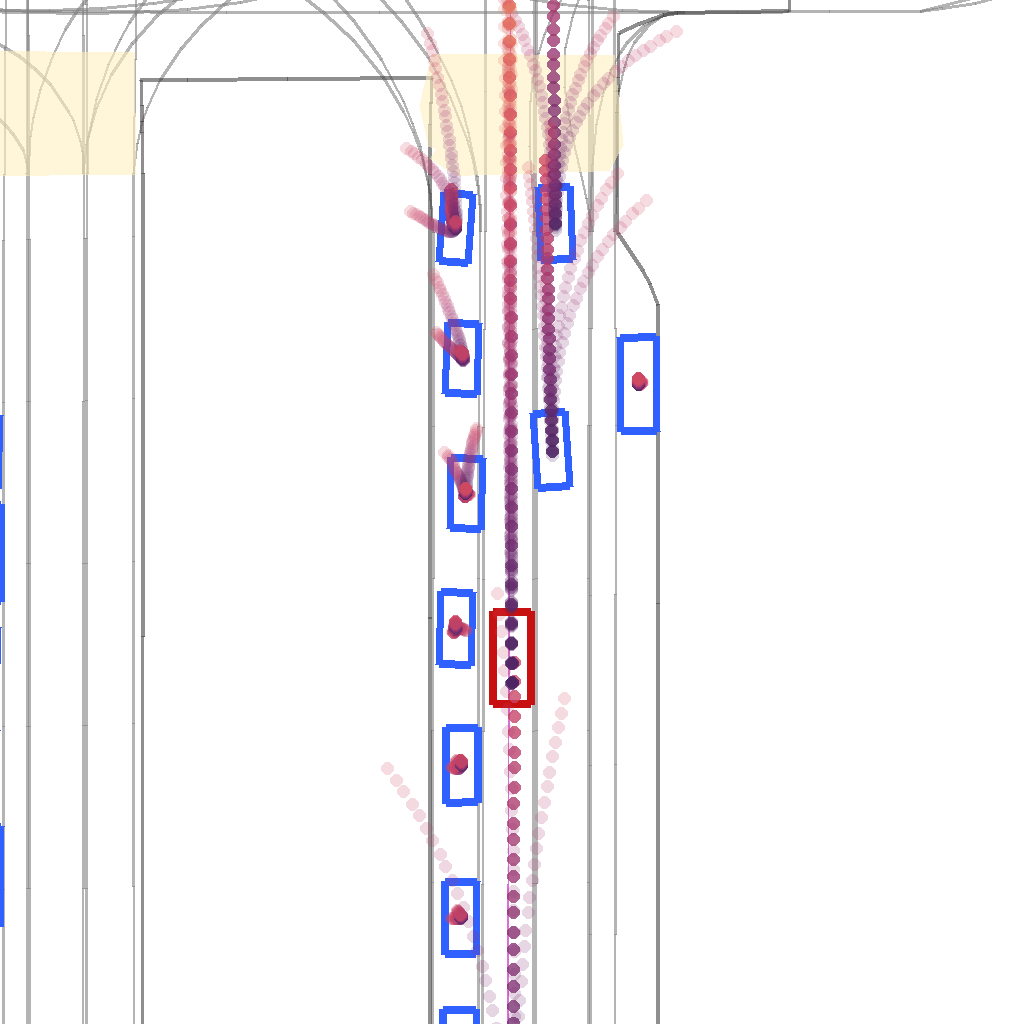} &
        \includegraphics[width=0.26\linewidth]{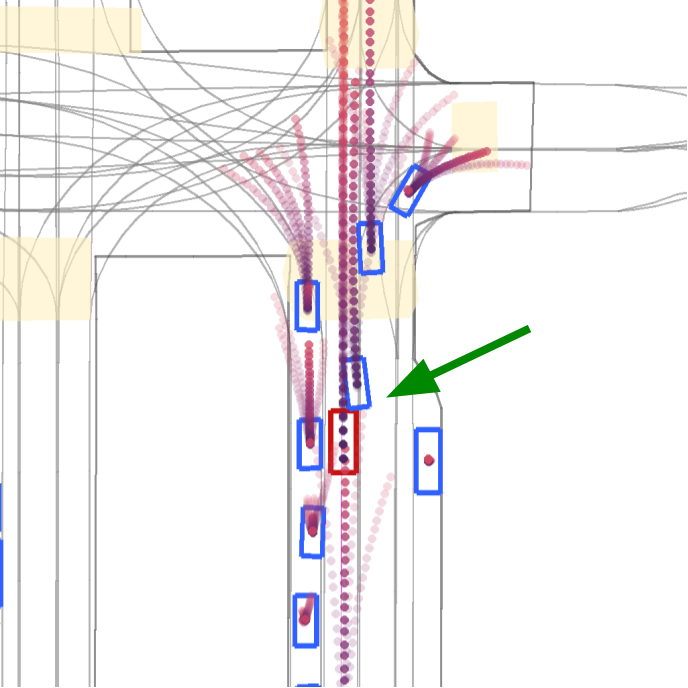} \\
        \rotatebox[origin=l]{90}{\hspace{3mm}\mincostcc} &
        \includegraphics[width=0.26\linewidth]{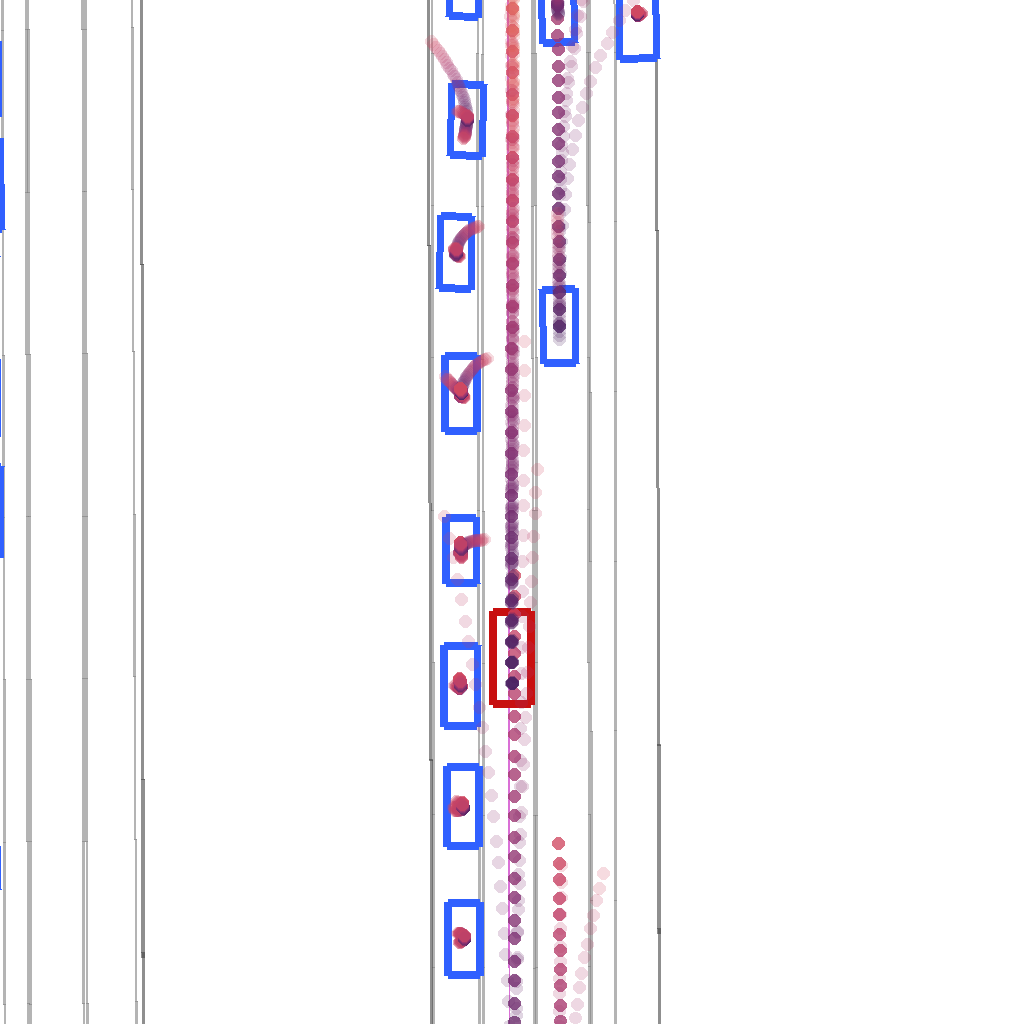} &
        \includegraphics[width=0.26\linewidth]{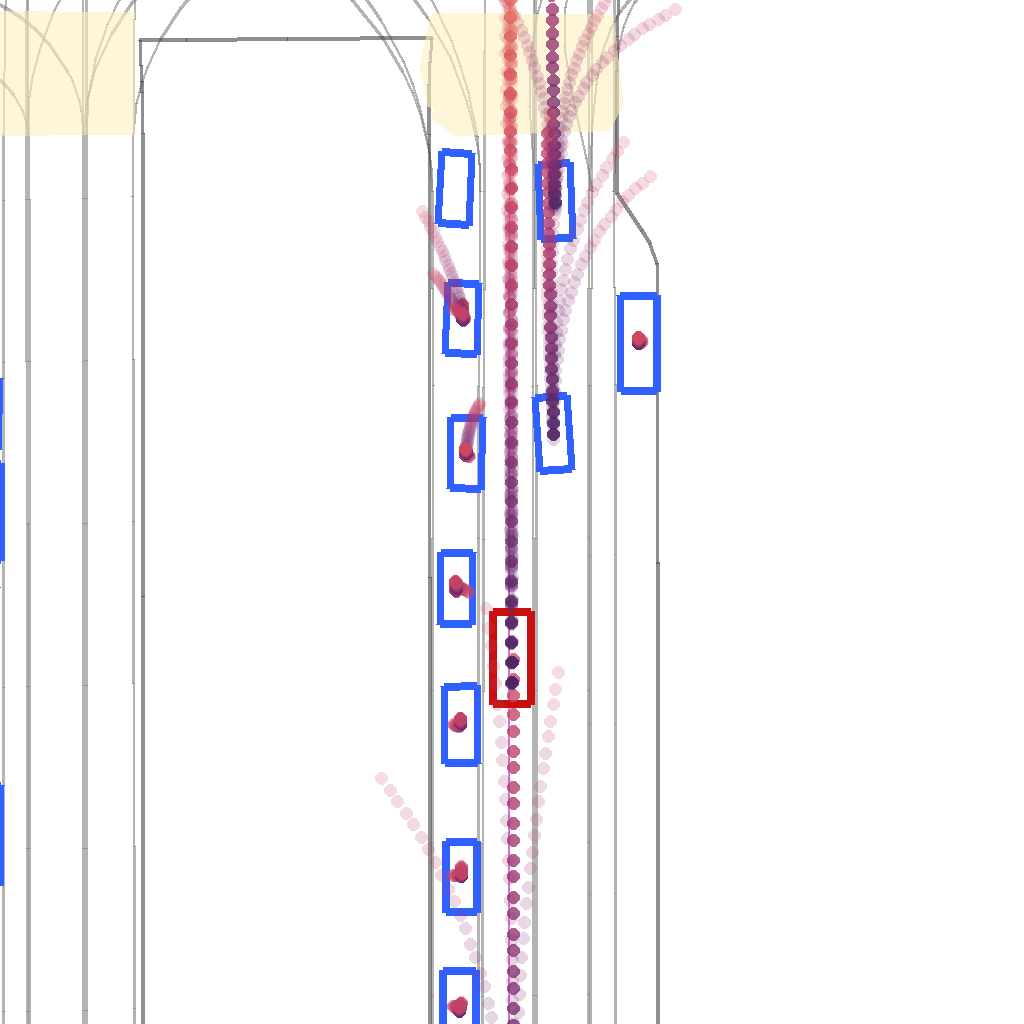} &
        \includegraphics[width=0.26\linewidth]{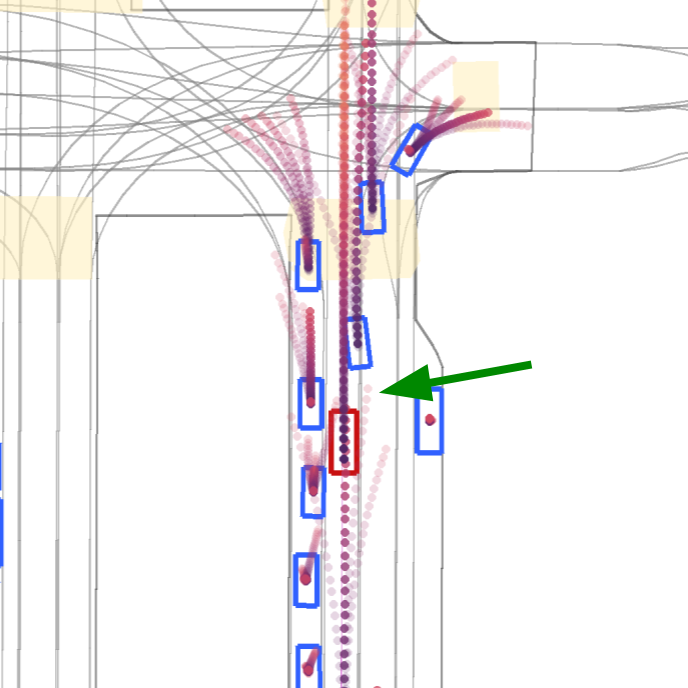} \\
        \midrule
        \rotatebox[origin=l]{90}{\hspace{4mm}\mincost} &
        \includegraphics[width=0.26\linewidth]{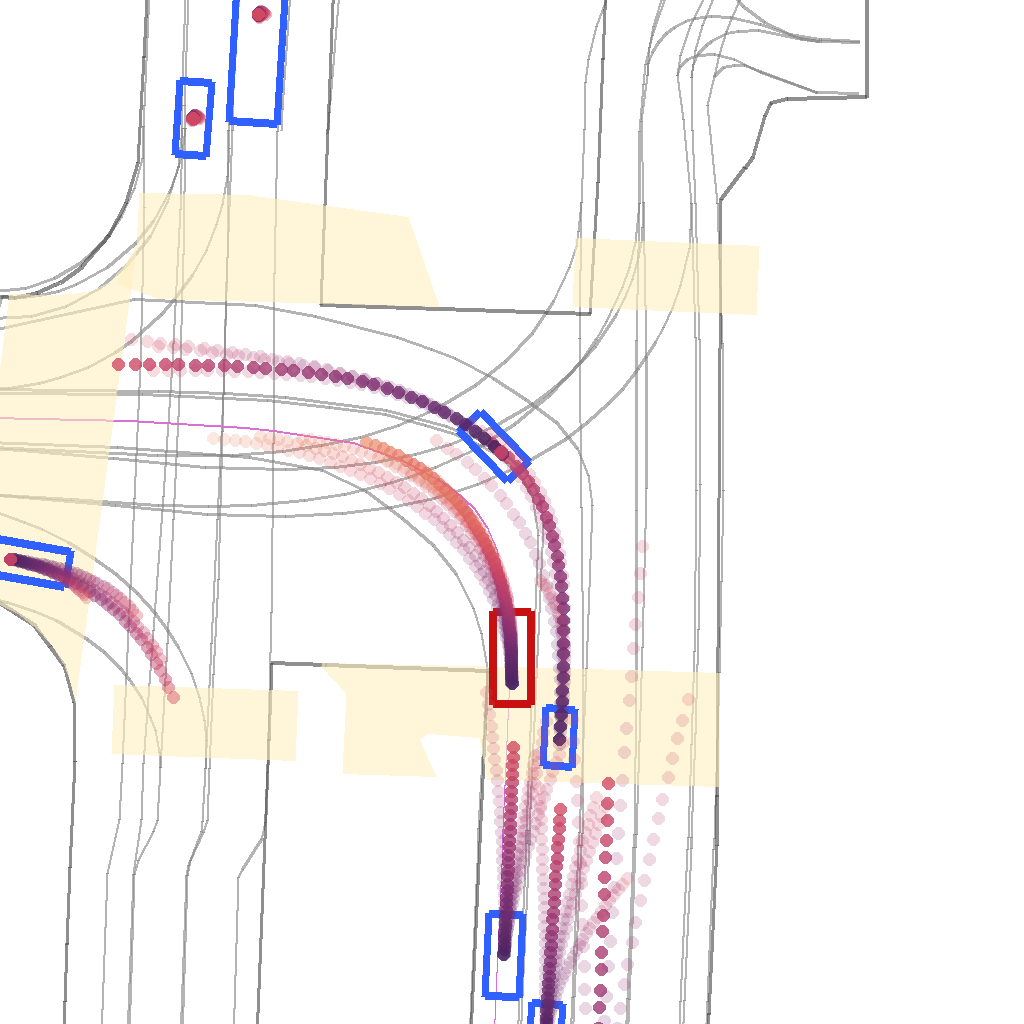} &
        \includegraphics[width=0.26\linewidth]{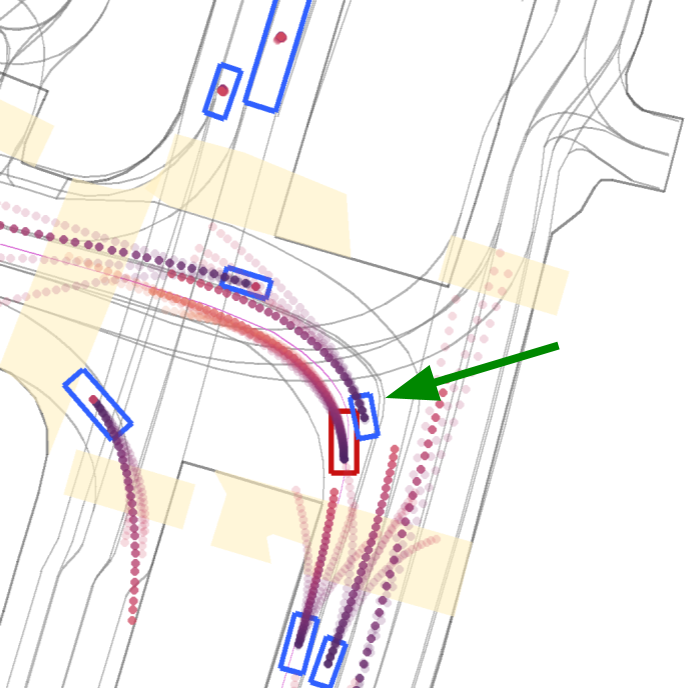} &
        \includegraphics[width=0.26\linewidth]{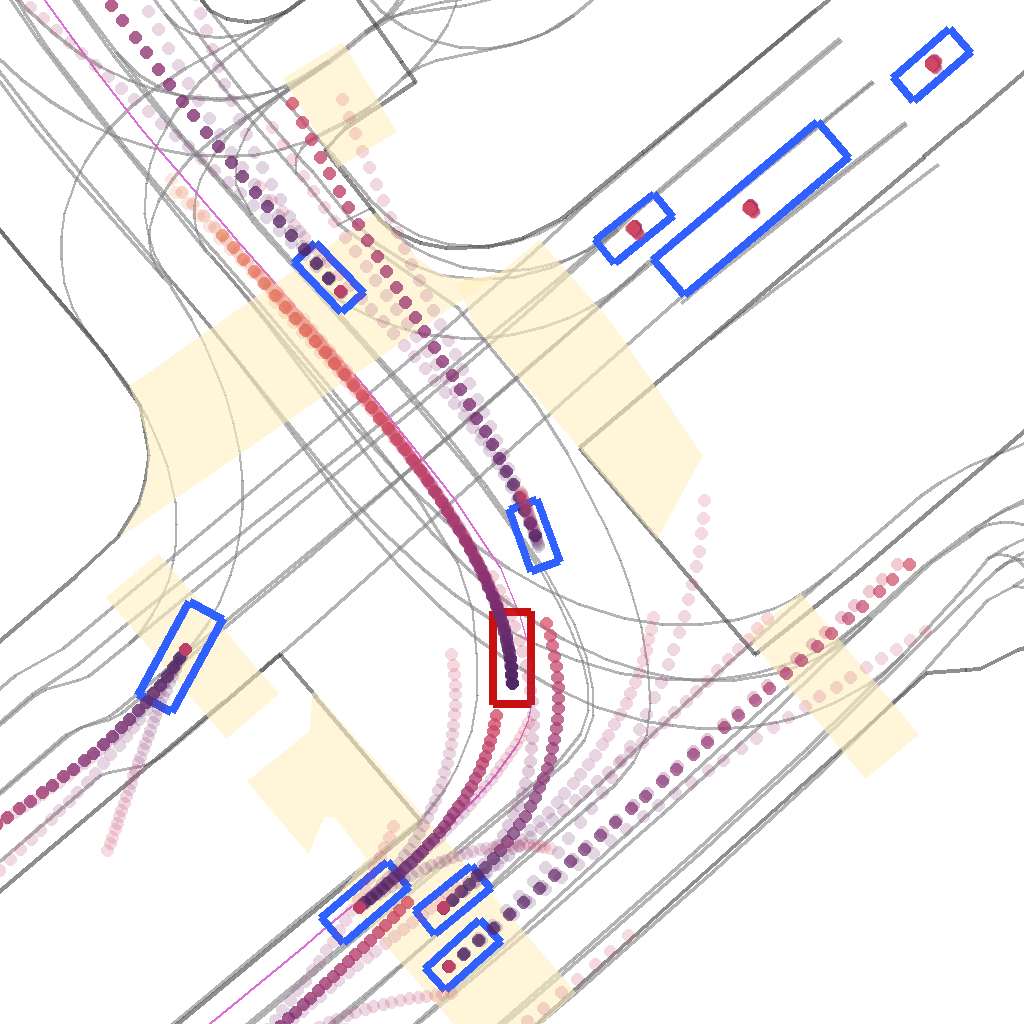} \\
        \rotatebox[origin=l]{90}{\hspace{3mm}\mincostcc} &
        \includegraphics[width=0.26\linewidth]{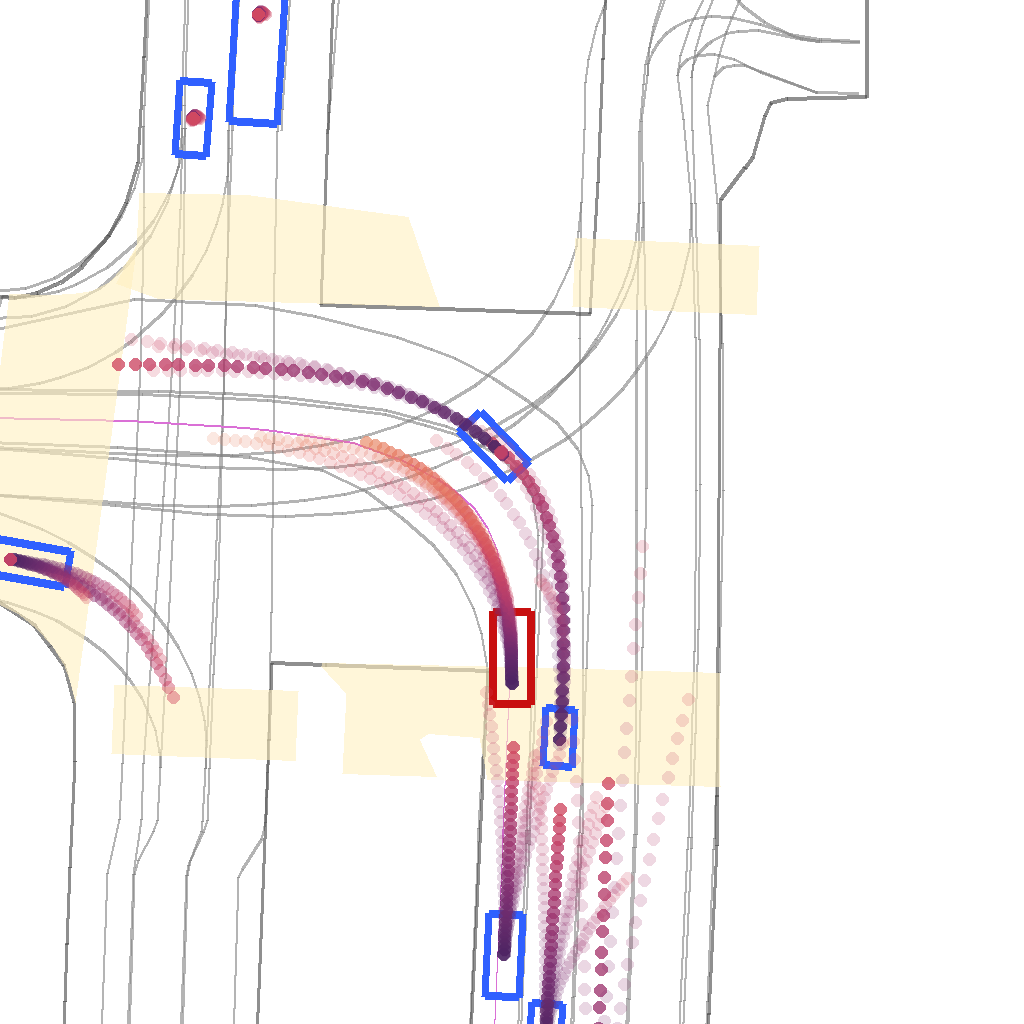} &
        \includegraphics[width=0.26\linewidth]{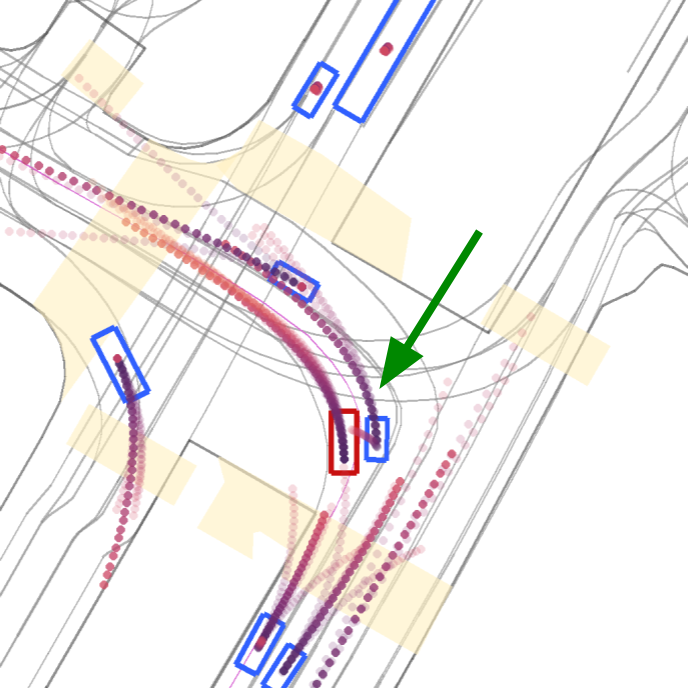} &
        \includegraphics[width=0.26\linewidth]{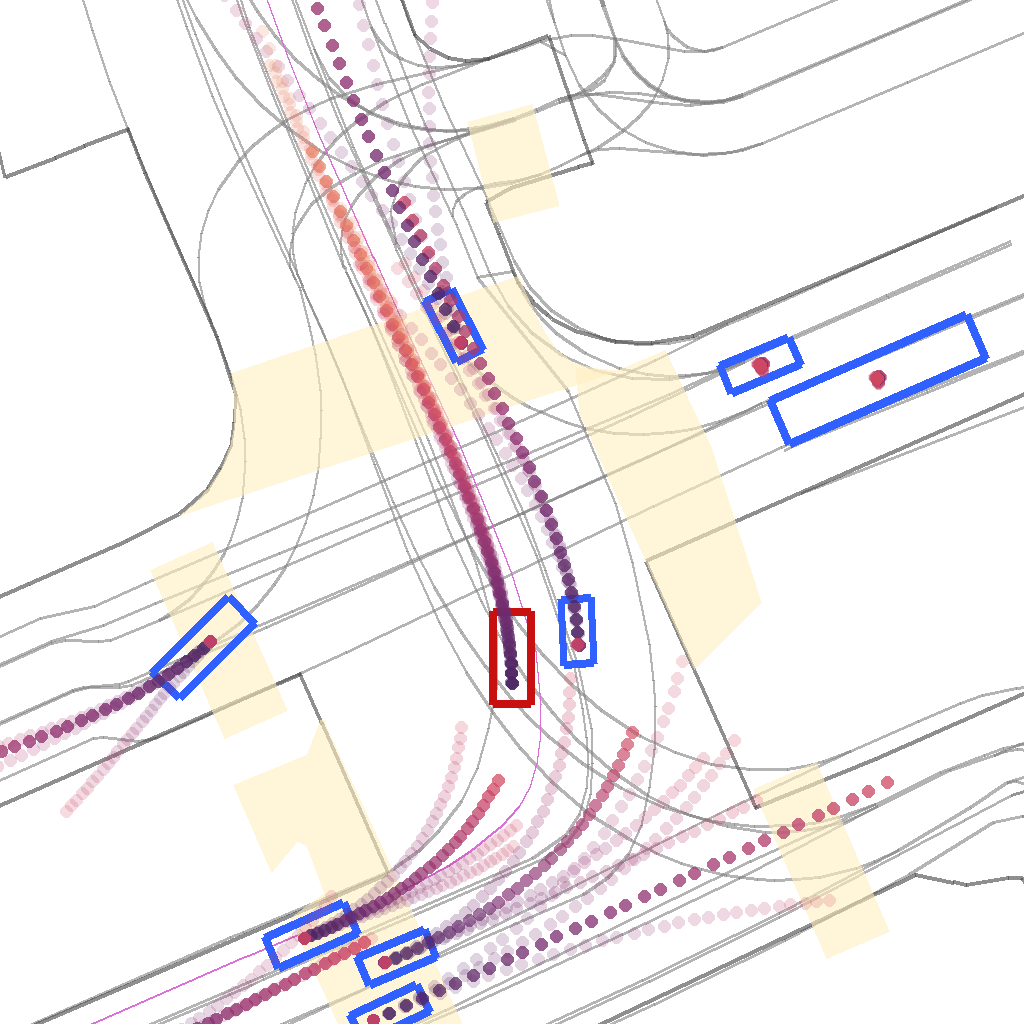} \\
        \multicolumn{4}{c}{
            \includegraphics[trim={0 1cm 0 1cm},clip,width=0.85\linewidth]{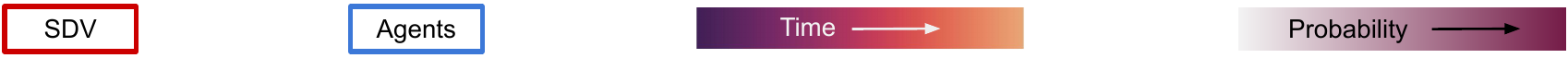}
        }
    \end{tabular}
\else
    \resizebox{\linewidth}{!}{
    \setlength{\tabcolsep}{3pt}
    \begin{tabular}{cccc c|c ccc}
        & \multicolumn{3}{c}{\resizebox{0.45\linewidth}{!}{\timearrow}} & \hspace{4pt} & ~
        & \multicolumn{3}{c}{\resizebox{0.45\linewidth}{!}{\timearrow}} \\
        \rotatebox[origin=l]{90}{\hspace{3mm}\footnotesize\mincost} &
        \includegraphics[width=0.15\linewidth]{images/comparison/img_11_01_MinCost.png} &
        \includegraphics[width=0.15\linewidth]{images/comparison/img_11_02_MinCost.png} &
        \includegraphics[width=0.15\linewidth]{images/comparison/img_11_03_MinCost_arrow.png} &  & &
        \includegraphics[width=0.15\linewidth]{images/comparison/img_12_01_MinCost.png} &
        \includegraphics[width=0.15\linewidth]{images/comparison/img_12_02_MinCost_arrow.png} &
        \includegraphics[width=0.15\linewidth]{images/comparison/img_12_03_MinCost.png} \\
        \rotatebox[origin=l]{90}{\hspace{1.5mm}\footnotesize\mincostcc} &
        \includegraphics[width=0.15\linewidth]{images/comparison/img_11_01_MinCostCC.png} &
        \includegraphics[width=0.15\linewidth]{images/comparison/img_11_02_MinCostCC.png} &
        \includegraphics[width=0.15\linewidth]{images/comparison/img_11_03_MinCostCC_arrow.png} & &  &
        \includegraphics[width=0.15\linewidth]{images/comparison/img_12_01_MinCostCC.png} &
        \includegraphics[width=0.15\linewidth]{images/comparison/img_12_02_MinCostCC_arrow.png} &
        \includegraphics[width=0.15\linewidth]{images/comparison/img_12_03_MinCostCC.png} \\
        \multicolumn{9}{c}{
            \includegraphics[trim={0 1cm 0 1cm},clip,width=0.85\linewidth]{figures/qualitative_legend.pdf}
        }
    \end{tabular}
    }
\fi
    \vspace{-3pt}
    \caption{Qualitative results of \ours, comparing \mincost and \mincostcc policy. 
    Combining \ours and \mincostcc improves the safety by reducing the estimated contacts with other road agents.
    }
    \label{fig:qualitative}
    \vspace{-11pt}
\end{figure}
\fi

\subsection{Results on Motion Prediction}
We also compare the quality of our agent predictions with the literature on the public Lyft Motion Prediction Dataset~\cite{houston2020one}. Results in terms of minADE and minFDE are reported in Table~\ref{tab:prediction}. The superscript value on each method represents the number of per-agent predicted trajectories.
As shown, our approach performs on par with or better than other methods and our unimodal baseline.

\subsection{Results from Road Testing}
Our model runs in real time on an SDV, integrated in the autonomous driving stack (10Hz). Thus, it can be deployed and tested on an SDV in the real world.

To validate our proposal in the real world, we first deployed and evaluated it on a private track. %
In particular, we successfully tested interactions with lead vehicles, including late braking at intersections and hard braking, and handling of traffic light intersections.
Then, we tested \ours on public roads letting it drive the SDV in full autonomy in Palo Alto.
As can be seen in the attached video,
our approach correctly handles challenging scenarios and safely drives on public roads. This includes complex situations and intersections where many other road agents are present.
The road deployment confirmed that the planner was comfortable and felt safe for the SDV passengers.

\subsection{Ablation study}\label{sec:ablation}
In this section we present results of an empirical ablation study performed in our efficient simulation.
The goal of the ablation study is to validate that the model learns a wide range of driving profiles,
and that all parts of the model are required for safe and comfortable driving.

\ifx\singlecol\undefined
    \begin{table*}[t]
    \centering
    \resizebox{\linewidth}{!}{
    \begin{tabular}[t]{cc}
    \begin{minipage}[b]{0.37\linewidth}
        \centering
        \captionof{table}{Comparison of prediction results on\\ Lyft Motion Prediction Dataset.}
        \label{tab:prediction}
        \resizebox{\linewidth}{!}{
            \begin{tabular}{l c c}
    \toprule
    Method &  minADE@3s & minFDE@3s \\
    \midrule
    SimNet~\cite{bergamini2021simnet}$^1$ & 0.71 & 1.38 \\
    Trajectron++~\cite{salzmann2020trajectronpp}$^{20}$ & 0.23 & 0.40 \\
    HAICU~\cite{ivanovic2021heterogeneous}$^{20}$ & 0.26 & 0.38 \\
    \midrule
    Ours$^1$ & 0.43 & 0.86 \\
    Ours$^{20}$ & 0.22 & 0.31 \\
    \bottomrule
\end{tabular}
        }
    \end{minipage}&\begin{minipage}[b]{0.63\linewidth}
        \centering
        \captionof{table}{Simulation ablation study comparing models trained with or without probability head and multiple experts
        (mean$\pm$std over 5 runs).
        }
        \label{tab:ablation}
        \resizebox{\linewidth}{!}{
            \begin{tabular}{c c c c c c c c}
\toprule
    \multirow{2}{*}{Prob. head} &
    \multicolumn{2}{c}{Multiple experts} &
    \multicolumn{2}{c}{\ilk} &
    \multicolumn{2}{c}{minADE@3s} \\ %
 & SDV & agents & \mincost & \mincostcc & SDV & agents \\
\midrule
\checkmark & \checkmark & \checkmark & 300 \textcolor{mygray}{$\pm$ 17.3} & 183 \textcolor{mygray}{$\pm$ 20.7} & 0.157 \textcolor{mygray}{$\pm$ 0.004} & 0.738 \textcolor{mygray}{$\pm$ 0.005} \\
\midrule
 & \checkmark & \checkmark & 571 \textcolor{mygray}{$\pm$ 175} & 359 $\pm$ \textcolor{mygray}{35.4} & 0.081 \textcolor{mygray}{$\pm$ 0.000} & 0.702 \textcolor{mygray}{$\pm$ 0.002} \\ 
\checkmark & \checkmark &  & 297 \textcolor{mygray}{$\pm$ 17.9} & 181 $\pm$ \textcolor{mygray}{23.1} & 0.156 \textcolor{mygray}{$\pm$ 0.004} & 0.838 \textcolor{mygray}{$\pm$ 0.007} \\
\checkmark &  & \checkmark & 298 \textcolor{mygray}{$\pm$ 18.9} & 298 \textcolor{mygray}{$\pm$ 18.9} & 0.316 \textcolor{mygray}{$\pm$ 0.003} & 0.726 \textcolor{mygray}{$\pm$ 0.007} \\ 
\bottomrule
\end{tabular}
        }
    \end{minipage}
    \end{tabular}
    }
    \end{table*}
\else
    \begin{table}[t]
    \centering
    \resizebox{\linewidth}{!}{
    \begin{tabular}[t]{cc}
    \begin{minipage}[b]{0.36\linewidth}
        \centering
        \captionof{table}{Comparison of prediction results on Lyft Motion Prediction Dataset.}
        \label{tab:prediction}
        \resizebox{\linewidth}{!}{
            \begin{tabular}{l c c}
    \toprule
    Method &  minADE@3s & minFDE@3s \\
    \midrule
    SimNet~\cite{bergamini2021simnet}$^1$ & 0.71 & 1.38 \\
    Trajectron++~\cite{salzmann2020trajectronpp}$^{20}$ & 0.23 & 0.40 \\
    HAICU~\cite{ivanovic2021heterogeneous}$^{20}$ & 0.26 & 0.38 \\
    \midrule
    Ours$^1$ & 0.43 & 0.86 \\
    Ours$^{20}$ & 0.22 & 0.31 \\
    \bottomrule
\end{tabular}
        }
    \end{minipage}&\begin{minipage}[b]{0.64\linewidth}
        \centering
        \captionof{table}{Simulation ablation study comparing models trained with or without probability head and multiple experts
        (mean $\pm$ std over 5 runs).
        }
        \label{tab:ablation}
        \resizebox{\linewidth}{!}{
            \begin{tabular}{c c c c c c c c}
\toprule
    \multirow{2}{*}{Prob. head} &
    \multicolumn{2}{c}{Multiple experts} &
    \multicolumn{2}{c}{\ilk} &
    \multicolumn{2}{c}{minADE@3s} \\ %
 & SDV & agents & \mincost & \mincostcc & SDV & agents \\
\midrule
\checkmark & \checkmark & \checkmark & 300 \textcolor{mygray}{$\pm$ 17.3} & 183 \textcolor{mygray}{$\pm$ 20.7} & 0.157 \textcolor{mygray}{$\pm$ 0.004} & 0.738 \textcolor{mygray}{$\pm$ 0.005} \\
\midrule
 & \checkmark & \checkmark & 571 \textcolor{mygray}{$\pm$ 175} & 359 $\pm$ \textcolor{mygray}{35.4} & 0.081 \textcolor{mygray}{$\pm$ 0.000} & 0.702 \textcolor{mygray}{$\pm$ 0.002} \\ 
\checkmark & \checkmark &  & 297 \textcolor{mygray}{$\pm$ 17.9} & 181 $\pm$ \textcolor{mygray}{23.1} & 0.156 \textcolor{mygray}{$\pm$ 0.004} & 0.838 \textcolor{mygray}{$\pm$ 0.007} \\
\checkmark &  & \checkmark & 298 \textcolor{mygray}{$\pm$ 18.9} & 298 \textcolor{mygray}{$\pm$ 18.9} & 0.316 \textcolor{mygray}{$\pm$ 0.003} & 0.726 \textcolor{mygray}{$\pm$ 0.007} \\ 
\bottomrule
\end{tabular}
        }
    \end{minipage}
    \end{tabular}
    }
    \end{table}
    \begin{figure}[t]
        \centering
        \subfigure{\resizebox{0.42\linewidth}{!}{\pgfplotstableread{ %
Label aggressiveness passiveness
MinCostCC                     32                     99
MinCost                       24                     81
~                              0                      0
10                          1920                      0
9                           1766                      3
8                           1665                     24
7                            460                    240
6                            171                   1601
5                            105                     42
4                             15                   1731
3                              7                   2326
2                              1                   2631
1                              0                   3000
}\testdata
\begin{tikzpicture}
\begin{axis}[
    xbar stacked,   %
    bar width=1.8ex,
    xmin=0,         %
    ytick=data,     %
    legend style={at={(axis cs:3200,-0.7)},anchor=south east},
    yticklabels from table={\testdata}{Label},  %
    height=6.5cm,
    xlabel={Number of metric violations},
    ylabel={~~~~~~~~~~~~~Action},
]
\addplot [fill=orange!80] table [x=aggressiveness, meta=Label,y expr=\coordindex] {\testdata};   %
\addplot [fill=teal!60] table [x=passiveness, meta=Label,y expr=\coordindex] {\testdata};
\legend{Aggressiveness, Passiveness}

\end{axis}

\end{tikzpicture}}\label{fig:ablation:profiles}}
        \hspace{0.07\linewidth}
        \subfigure{\resizebox{0.385\linewidth}{!}{\begin{tikzpicture}
\begin{axis}[
    title={},
    xlabel={Dataset size [\%]},
    ylabel={\ilk},
    xmin=10, xmax=100,
    ymin=0, ymax=900,
    xtick={10,25,50,100},
    legend pos=north east,
    legend cell align={left},
    ymajorgrids=true,
    grid style=dashed,
    height=6.5cm
]

\addplot[
    color=RedOrange, %
    line width=0.5mm,
    mark=square,
    ]
    coordinates {
    (10, 837.5191650390625)
    (25, 400.2679443359375)
    (50, 283.8168640136719)
    (100, 278.5482482910156)
    };
    \addlegendentry{\mincost}

\addplot[
    color=Cerulean, %
    line width=0.4mm,
    mark=square,
    ]
    coordinates {
    (10, 479.7998962402344)
    (25, 268.472900390625)
    (50, 197.9602813720703)
    (100, 168.9991455078125)
    };
    \addlegendentry{\mincostcc}
    
\addplot[
    color=Red, %
    line width=0.6mm,
    dashdotted,
    ]
    coordinates {
    (10, 104)
    (100, 104)
    };
    \addlegendentry{log-replay}

\end{axis}

\end{tikzpicture}}\label{fig:ablation:ilk}}
        \caption{
        (a) We evaluate a single model using fixed expert selection policy.
        The model learns a diverse set of driving profiles, ranging from aggressive 
        to passive. 
        When augmenting \mincost policy with collision check (\mincostcc),
        we get a safe and comfortable driving policy.
        (b) We train our model using different subsets of training data,
        and evaluate them using \mincost and \mincostcc policies.
        \mincostcc policy requires fewer
        data samples to achieve performance of \mincost.
        Moreover, it keeps improving with more data, while \mincost performance plateaus.
        }
        \label{fig:ablation}
    \end{figure}
\fi

We report 
significant events per 1k miles (e.g. estimated contacts, close calls, divergence from route) 
and minADE metrics on the validation set.
We run the planner in log replay mode to obtain baseline \ilk.
Since the dataset was created using an automated perception system,
SDV and agent bounding boxes might overlap due to localization errors,
resulting in baseline \ilk close to 100.

\ablationsubsection{Evaluating driving policies separately.}
In this study, we take a single model predicting 10 SDV experts and 5 agent experts,
and perform simulation using fixed SDV expert selection, ignoring probabilities and collision checks.
Results can be seen in 
Fig.~\ref{fig:ablation:profiles}.
We show 
aggressiveness and passiveness metrics,
indicating whether the simulated SDV advances or lags behind compared to logged position.
As can be seen, SDV experts cover a wide range of behaviors,
from fast aggressive to slow passive driving.
Learning expert selection gives a safe and comfortable policy.

\ablationsubsection{Planning policy.}
To validate that \mincostcc policy shows fewer estimated contacts than \mincost, and that the model improves with more training data,
we train multiple models using 10\%, 25\%, 50\% and 100\% of training data,
and evaluate using both policies.
Results are in 
Fig.~\ref{fig:ablation:ilk}.
Both policies improve with more data samples up to 50\%, after which only \mincostcc improves.
\mincostcc requires significantly fewer training samples to achieve similar performance and can steadily improve adding more training data.

\ifx\singlecol\undefined
\begin{figure}[t]
    \centering
    \resizebox{0.71\linewidth}{!}{\pgfplotstableread{ %
Label aggressiveness passiveness
MinCostCC                     32                     99
MinCost                       24                     81
~                              0                      0
10                          1920                      0
9                           1766                      3
8                           1665                     24
7                            460                    240
6                            171                   1601
5                            105                     42
4                             15                   1731
3                              7                   2326
2                              1                   2631
1                              0                   3000
}\testdata
\begin{tikzpicture}
\begin{axis}[
    xbar stacked,   %
    bar width=1.8ex,
    xmin=0,         %
    ytick=data,     %
    legend style={at={(axis cs:3200,-0.7)},anchor=south east},
    yticklabels from table={\testdata}{Label},  %
    height=6.5cm,
    xlabel={Number of metric violations},
    ylabel={~~~~~~~~~~~~~Action},
]
\addplot [fill=orange!80] table [x=aggressiveness, meta=Label,y expr=\coordindex] {\testdata};   %
\addplot [fill=teal!60] table [x=passiveness, meta=Label,y expr=\coordindex] {\testdata};
\legend{Aggressiveness, Passiveness}

\end{axis}

\end{tikzpicture}}
    \caption{
    We evaluate a single model using fixed selection policy.
    The model learns a diverse set of driving profiles, ranging from aggressive 
    to passive. 
    When augmenting \mincost policy with collision check (\mincostcc),
    we get a safe and comfortable driving policy.
    }
    \label{fig:ablation:profiles}
    \vspace{-10pt}
\end{figure}
\else
\fi

\ablationsubsection{Disabling probability distribution learning.}
We train a baseline model and a model without the probability head in Eq.~\eqref{eq:assignment_cost} and \eqref{eq:main_loss}, setting $\lambda_\text{SDV}=\mu_\text{SDV}=0$.
Results can be found in Table~\ref{tab:ablation}.
The model without the probability head shows significantly lower SDV minADE,
but shows poor driving policy with many estimated contacts, as expected.

\ablationsubsection{Unimodal predictions.}
To validate the importance of having multiple experts for agent prediction
and SDV planning,
we train a model setting the number of
agent trajectories $M$ to 1 and another setting the number of SDV plans $N$ to 1.
Results can be found in Table~\ref{tab:ablation}.
In the case of agent trajectories, setting $M=1$ leads to higher agent minADE, confirming that using multiple agent trajectories is beneficial for the predictions, even though it does not significantly affect the \ilk with collision check policy.
In the case of SDV plans, setting $N=1$ leads to substantially more \ilk and higher SDV minADE, confirming that modelling multiple SDV trajectories is beneficial for both the policies.

\ifx\singlecol\undefined
\begin{figure}[t]
    \centering
    \resizebox{0.65\linewidth}{!}{\begin{tikzpicture}
\begin{axis}[
    title={},
    xlabel={Dataset size [\%]},
    ylabel={\ilk},
    xmin=10, xmax=100,
    ymin=0, ymax=900,
    xtick={10,25,50,100},
    legend pos=north east,
    legend cell align={left},
    ymajorgrids=true,
    grid style=dashed,
    height=6.5cm
]

\addplot[
    color=RedOrange, %
    line width=0.5mm,
    mark=square,
    ]
    coordinates {
    (10, 837.5191650390625)
    (25, 400.2679443359375)
    (50, 283.8168640136719)
    (100, 278.5482482910156)
    };
    \addlegendentry{\mincost}

\addplot[
    color=Cerulean, %
    line width=0.4mm,
    mark=square,
    ]
    coordinates {
    (10, 479.7998962402344)
    (25, 268.472900390625)
    (50, 197.9602813720703)
    (100, 168.9991455078125)
    };
    \addlegendentry{\mincostcc}
    
\addplot[
    color=Red, %
    line width=0.6mm,
    dashdotted,
    ]
    coordinates {
    (10, 104)
    (100, 104)
    };
    \addlegendentry{log-replay}

\end{axis}

\end{tikzpicture}}
    \caption{
    We train our model using different subsets of training data,
    and evaluate them using \mincost and \mincostcc policies.
    \mincostcc policy requires fewer
    data samples to achieve performance of \mincost.
    Moreover, it keeps improving with more data, while \mincost performance plateaus.
    }
    \label{fig:ablation:ilk}
    \vspace{-10pt}
\end{figure}
\else
\begin{figure}
    \centering
    \footnotesize
    
    \vspace{-3pt}
    \caption{Qualitative results of \ours, comparing \mincost and \mincostcc policy. 
    Combining \ours and \mincostcc improves the safety by reducing the estimated contacts with other road agents.
    (a) In the first simulated scene, we see the SDV colliding with the agent in the third shown frame when using \mincost whereas the SDV slows down using \mincostcc. (b) \mincost policy causes a collision with the turning agent in the second frame whereas \mincostcc shows safe distance between SDV and agent.
    }
    \label{fig:qualitative}
\end{figure}

\begin{figure}
    \centering
    \footnotesize
    \ifx\singlecol\undefined\
\else
    \resizebox{\linewidth}{!}{
    \setlength{\tabcolsep}{3pt}
    \begin{tabular}{ccc c|c ccc}
        \multicolumn{3}{c}{\resizebox{0.45\linewidth}{!}{\timearrow}} & \hspace{4pt} & ~
        & \multicolumn{3}{c}{\resizebox{0.45\linewidth}{!}{\timearrow}} \\
        \includegraphics[width=0.15\linewidth]{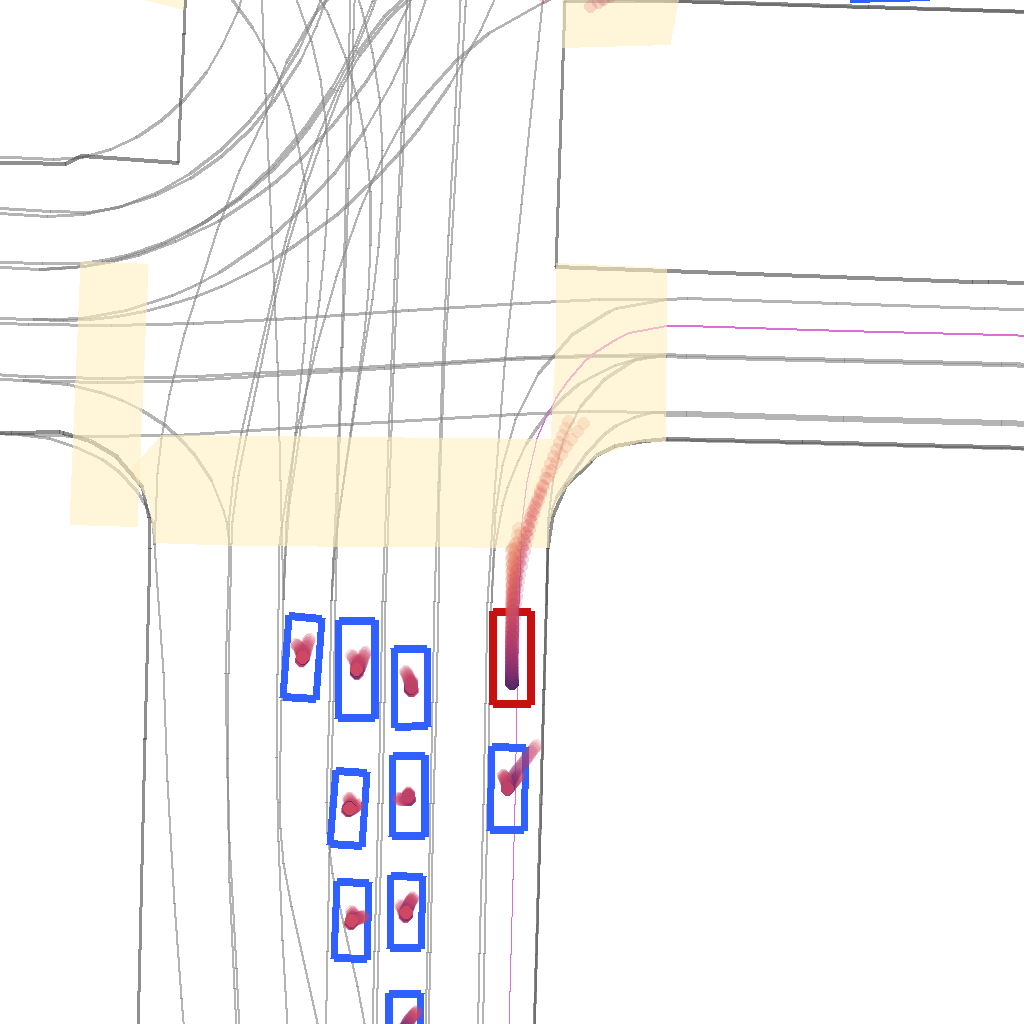} &
        \includegraphics[width=0.15\linewidth]{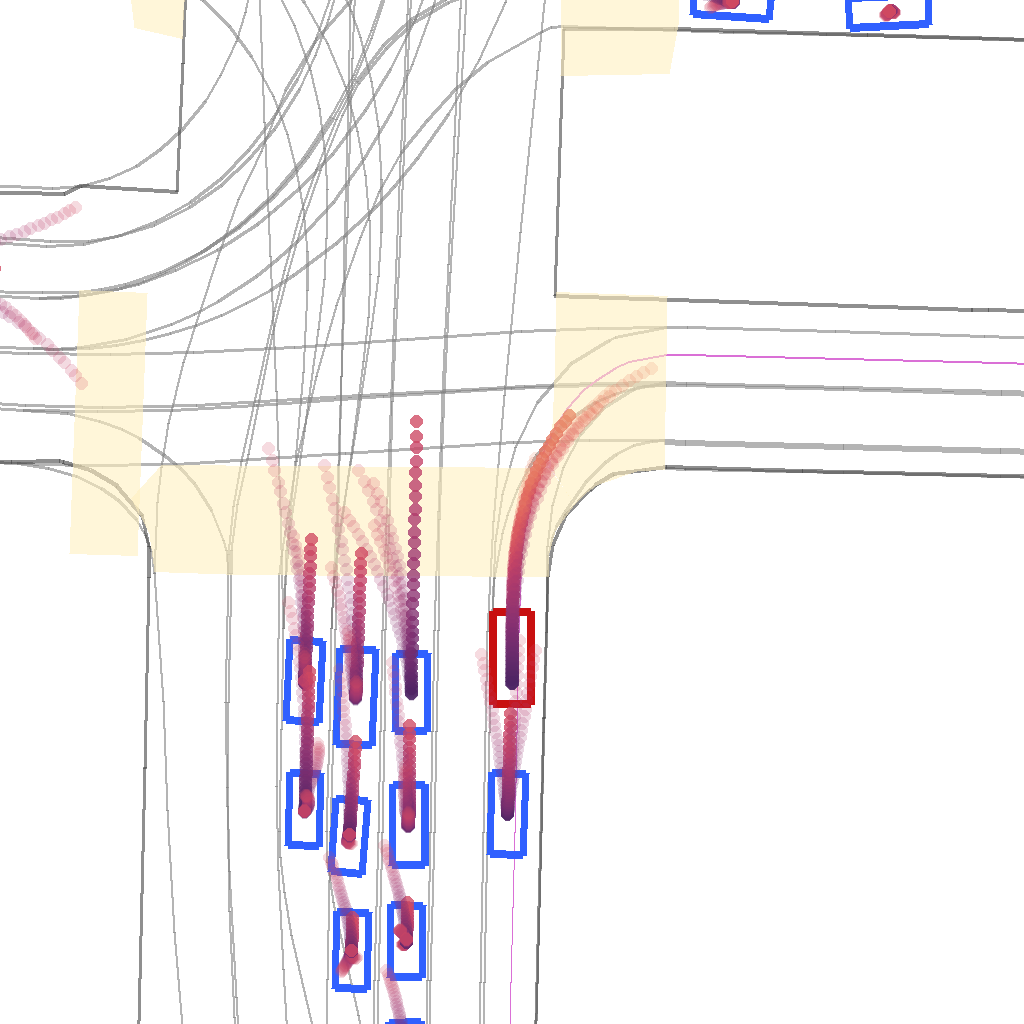} &
        \includegraphics[width=0.15\linewidth]{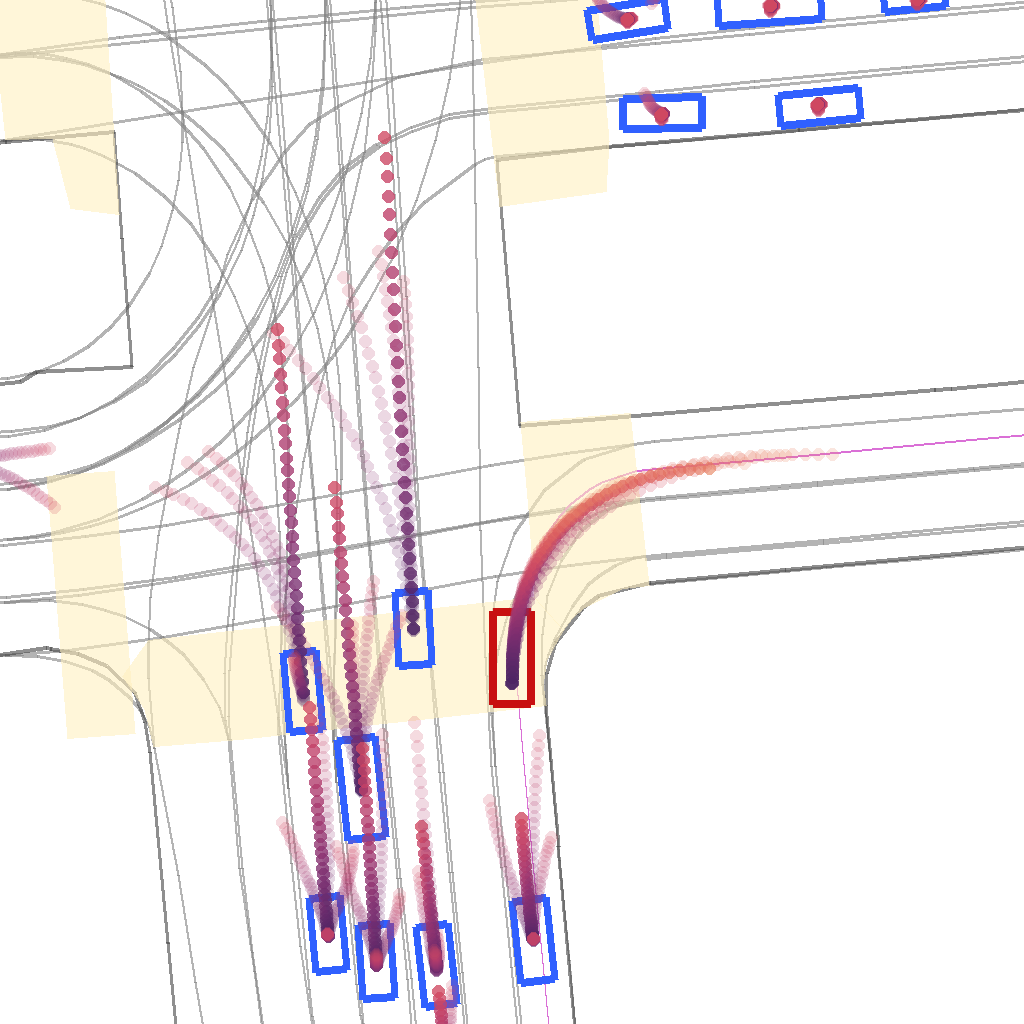} &  & &
        \includegraphics[width=0.15\linewidth]{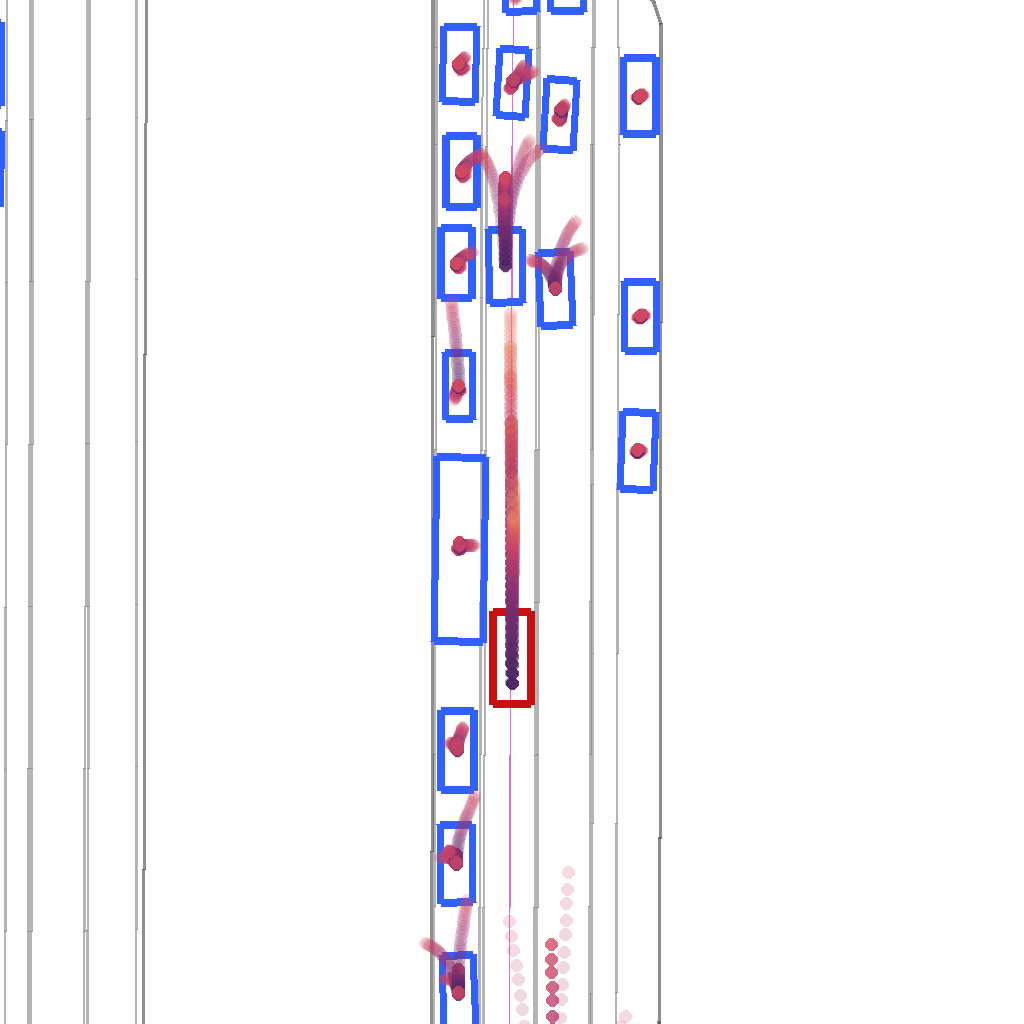} &
        \includegraphics[width=0.15\linewidth]{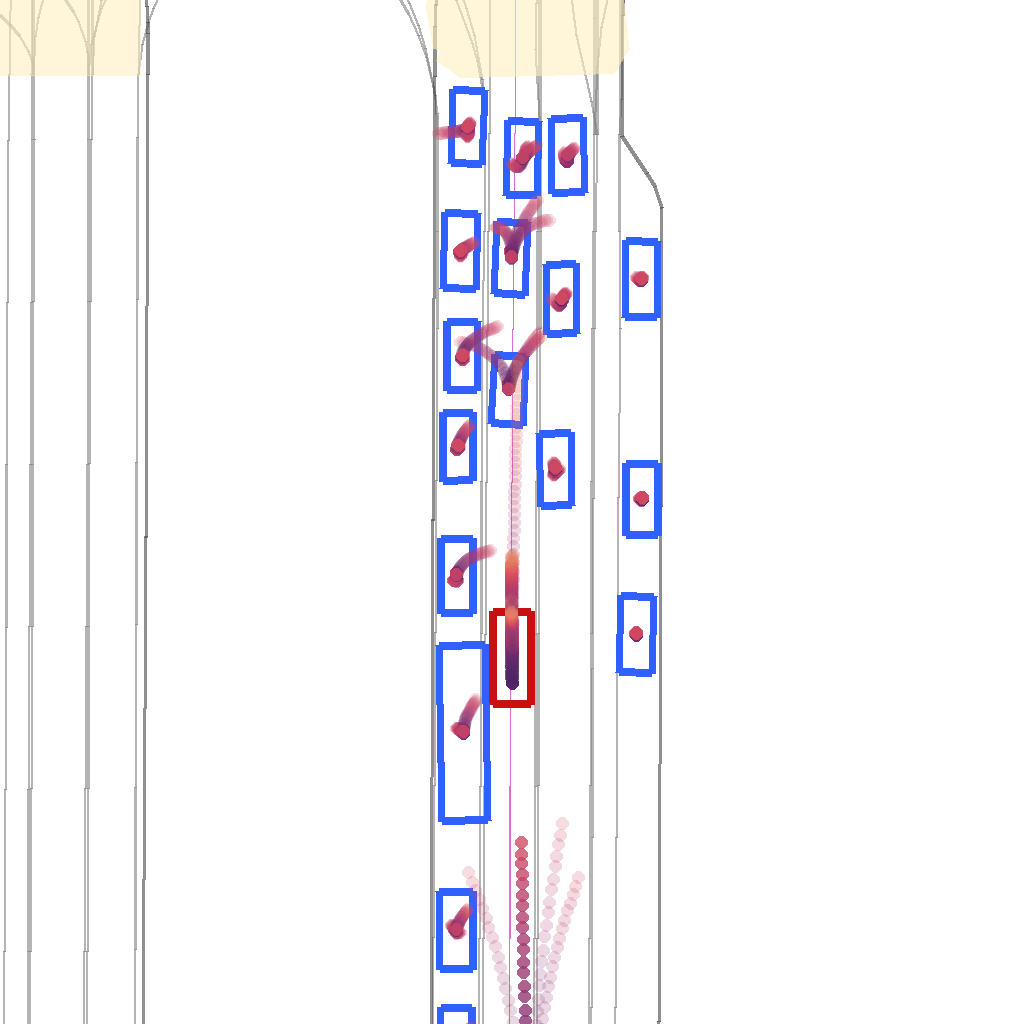} &
        \includegraphics[width=0.15\linewidth]{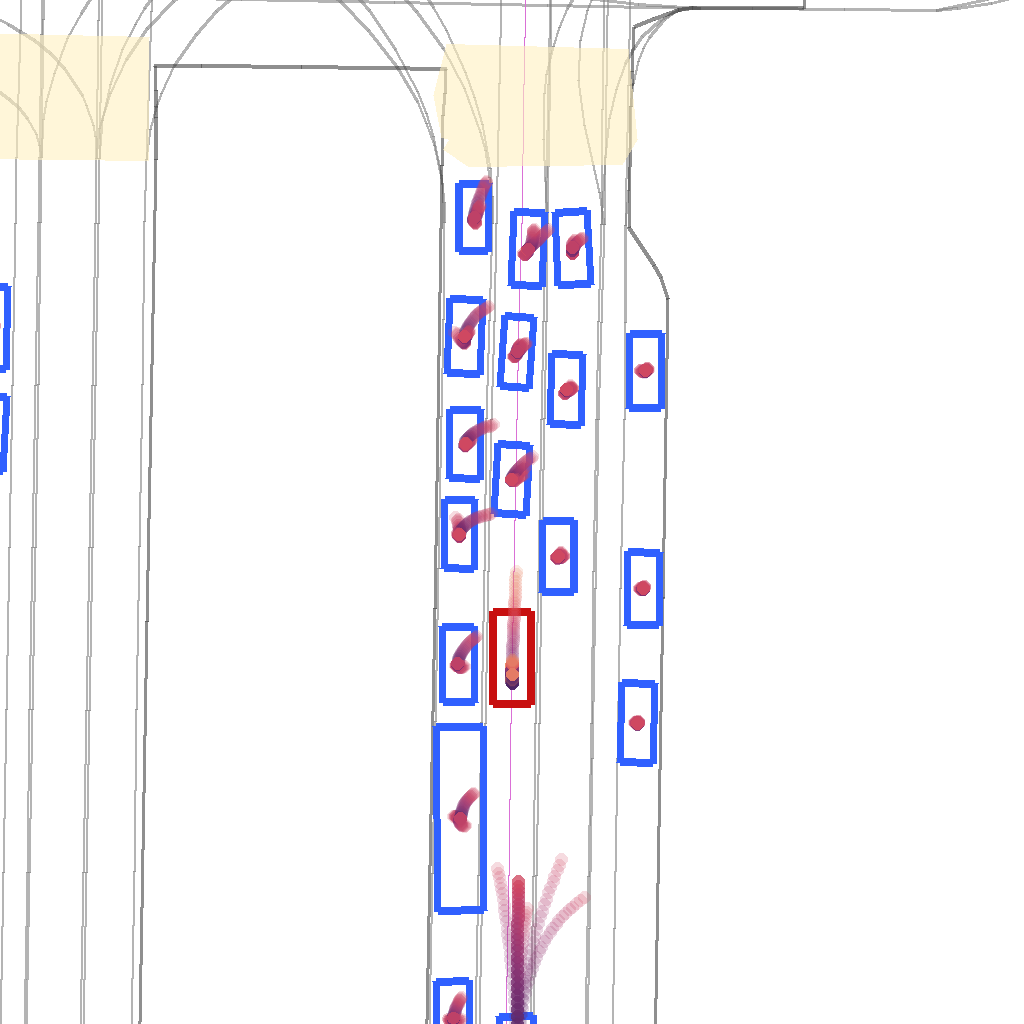} \\
        \includegraphics[width=0.15\linewidth]{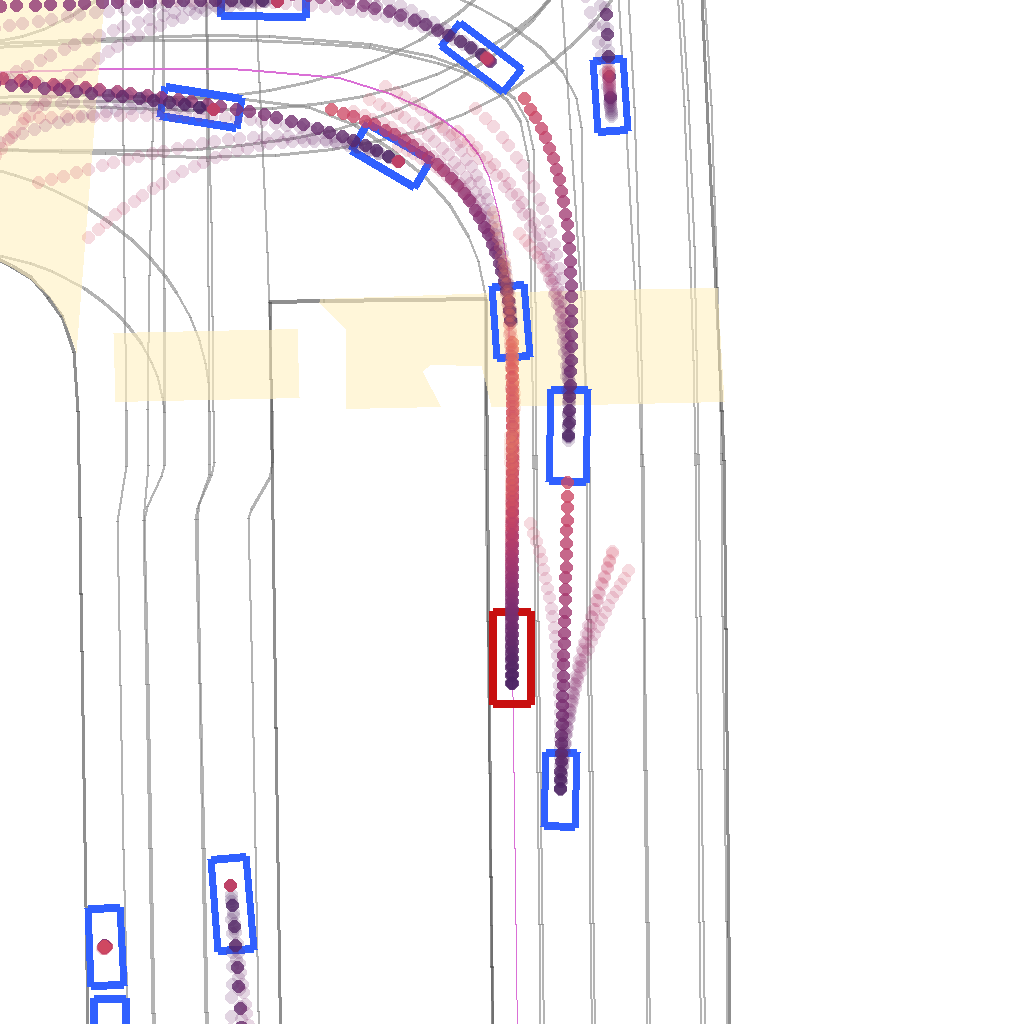} &
        \includegraphics[width=0.15\linewidth]{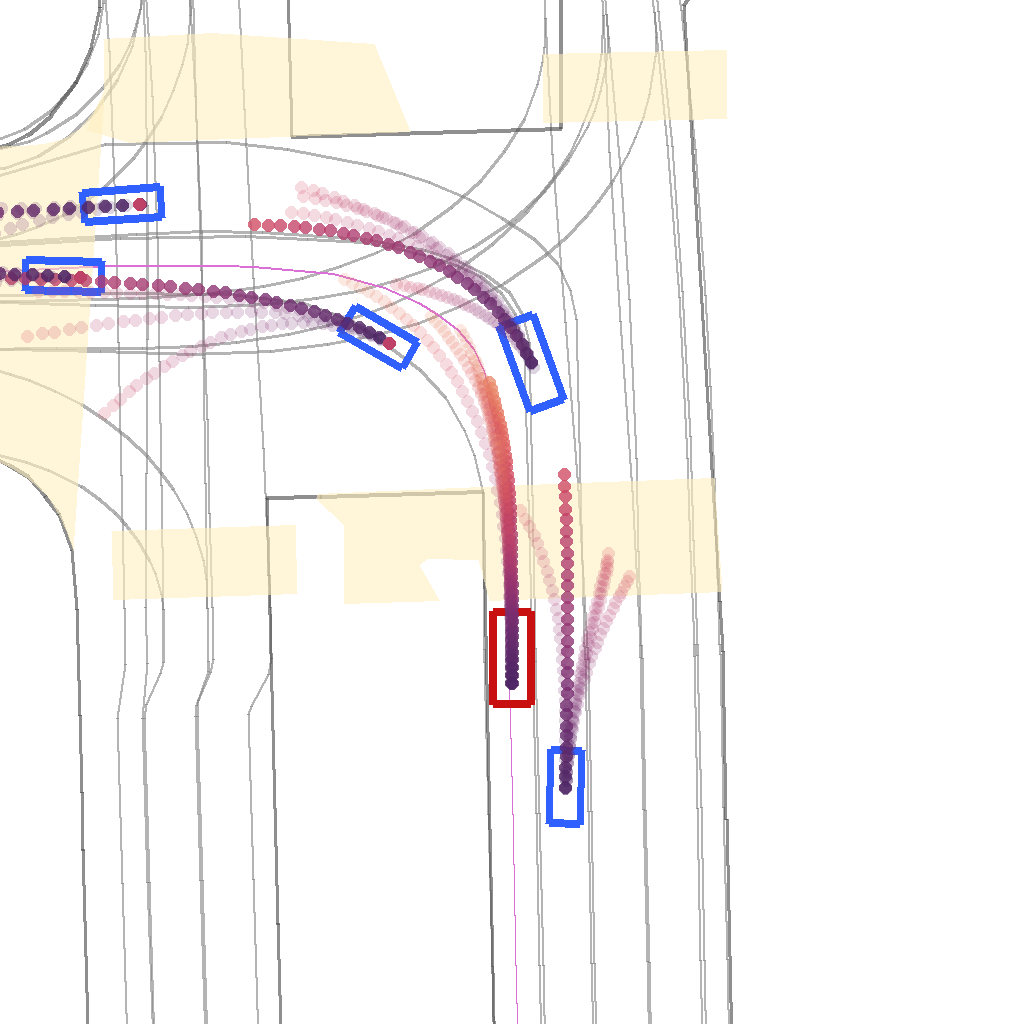} &
        \includegraphics[width=0.15\linewidth]{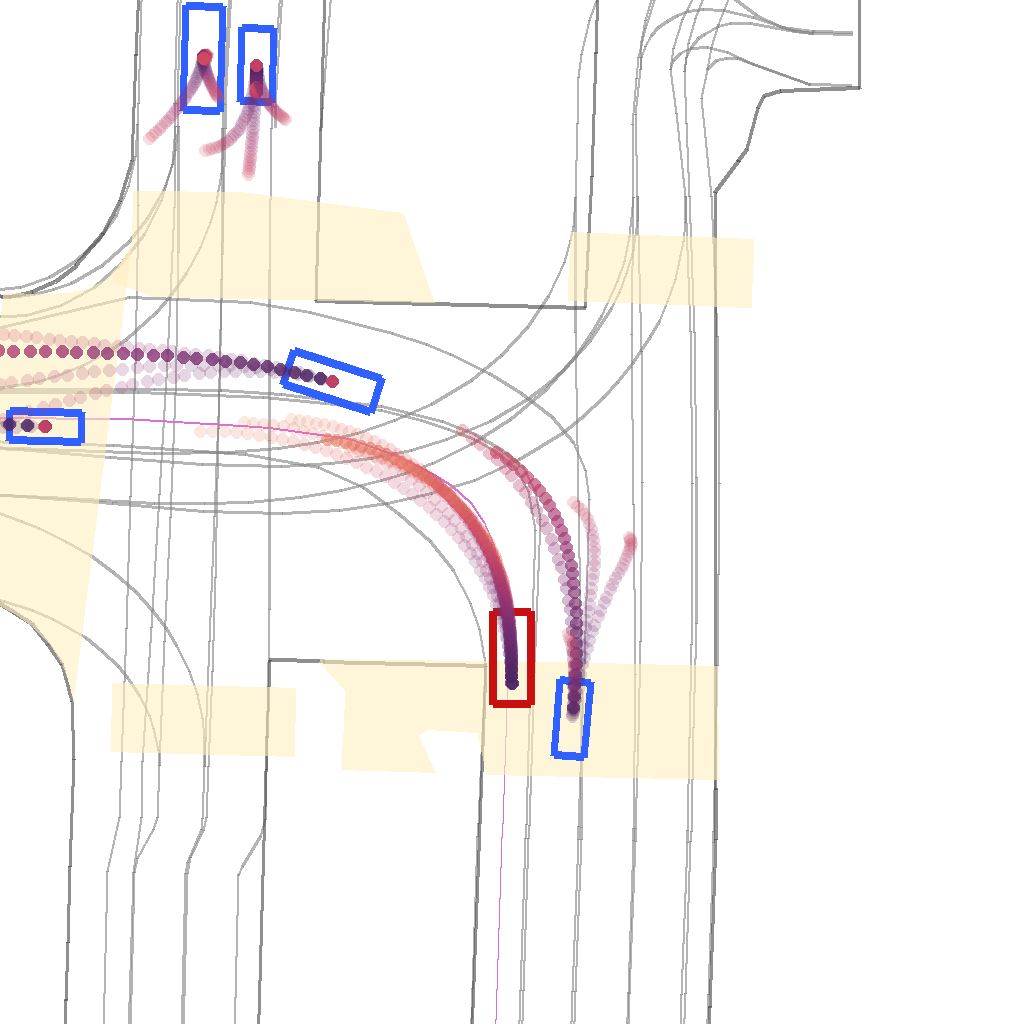} & &  &
        \includegraphics[width=0.15\linewidth]{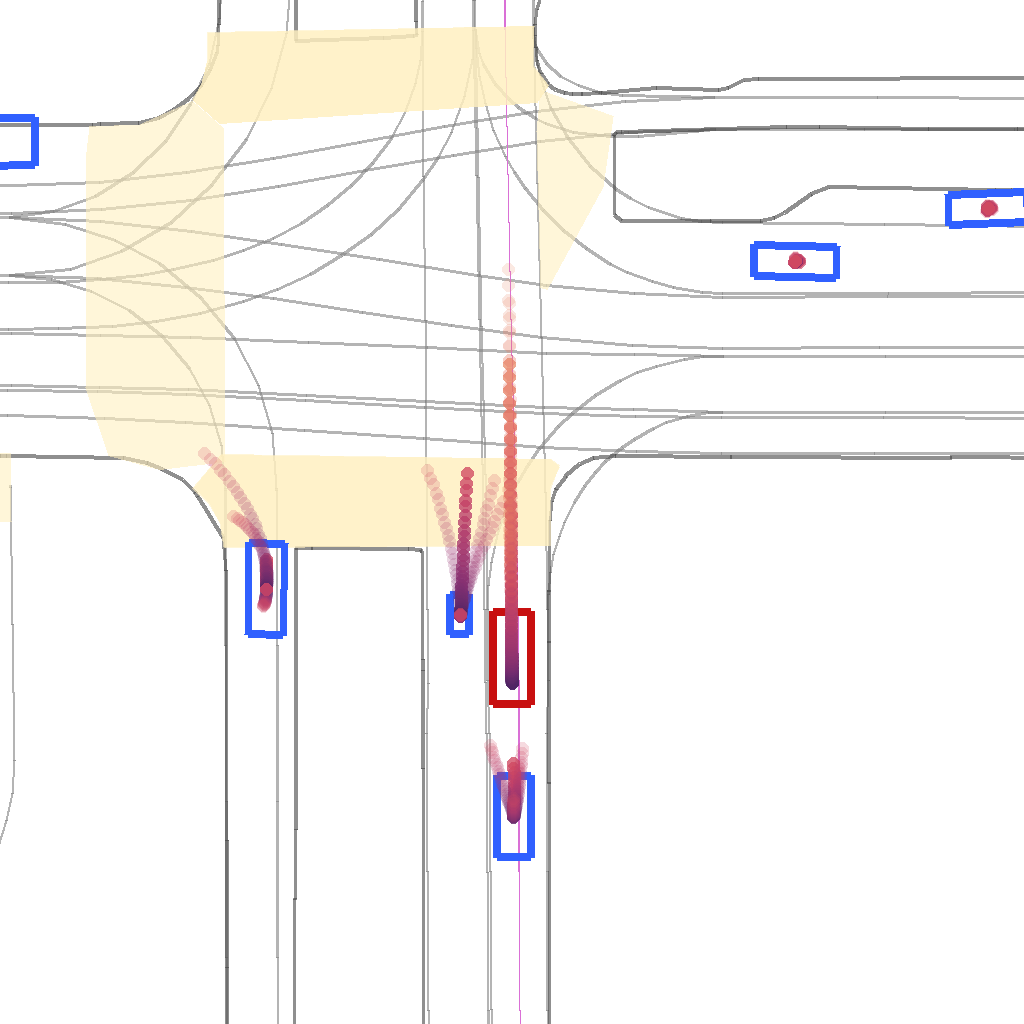} &
        \includegraphics[width=0.15\linewidth]{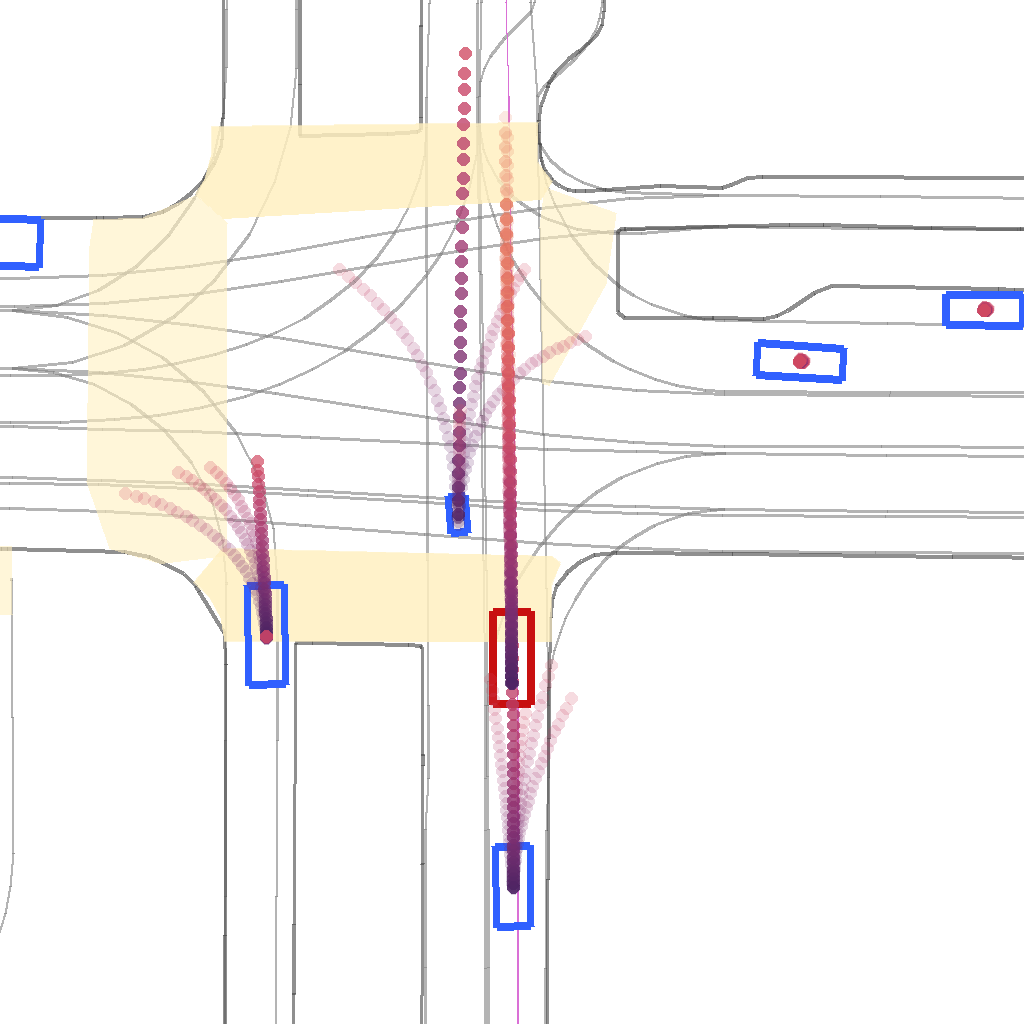} &
        \includegraphics[width=0.15\linewidth]{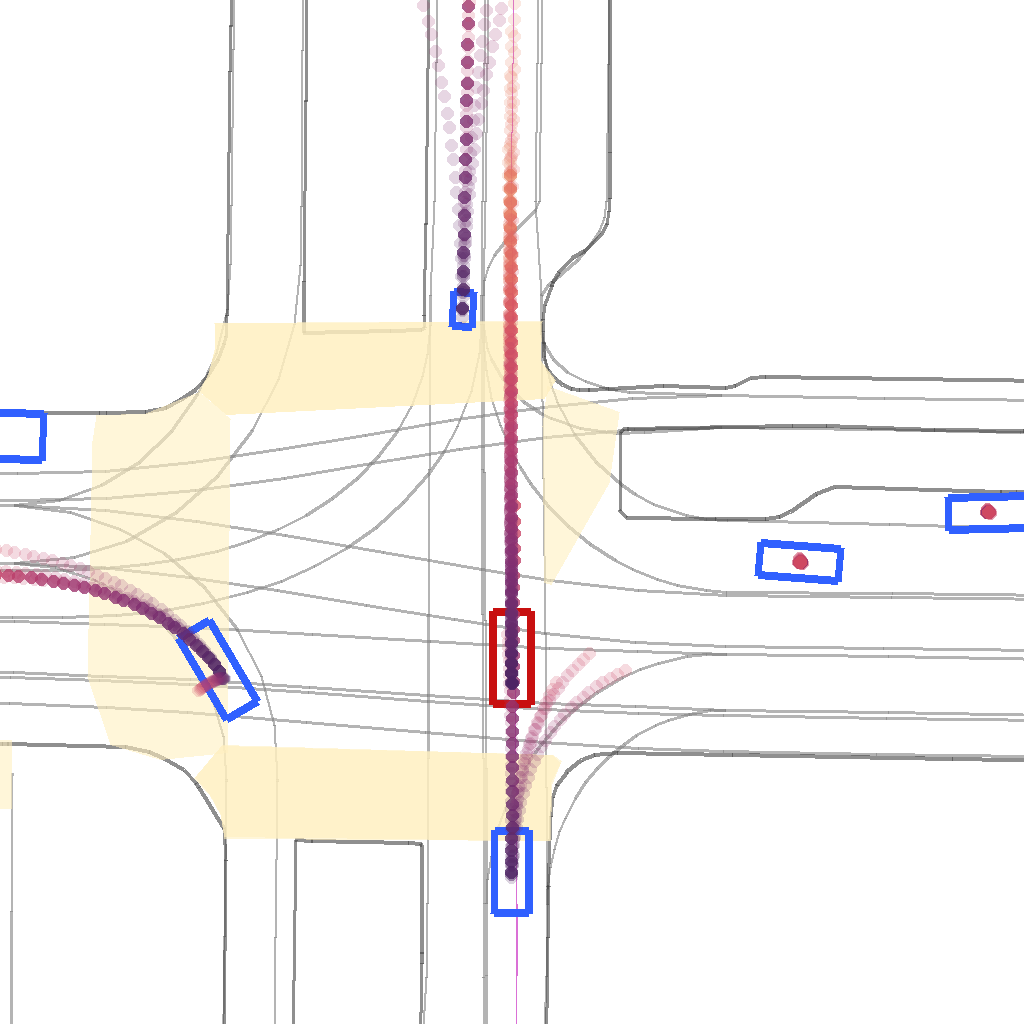} \\
        \multicolumn{8}{c}{
            \includegraphics[trim={0 1cm 0 1cm},clip,width=0.85\linewidth]{figures/qualitative_legend.pdf}
        }
    \end{tabular}
    }
\fi
    \vspace{-3pt}
    \caption{Additional qualitative results of \ours.
    (i) In the top left sample, the SDV makes a right turn. The planned trajectory curves as expected at the intersection. (ii) The top right sample shows a scene with dense traffic. The SDV comfortably comes to a stop while the other agents around it are stopped. (iii) In the bottom left sample, the SDV approaches a left turn. The agents around it also have accurate predictions. (iv) In the bottom right sample, the SDV proceeds at a green traffic light. Different speed profiles can be seen. The agent to the left of SDV (in the leftmost lane) changes its most probable mode from going straight to left turn as it continues through the intersection.
    }
    \label{fig:qualitative_additional}
\end{figure}
\fi

\section{Discussion}\label{sec:limitations}
In this paper we presented \ours, a unified neural model for prediction and planning that learns a distribution over future trajectories, corresponding to different driving profiles. %
Leveraging the predicted trajectory distribution, the safety of the planner is improved by penalizing SDV plans that could lead to collisions with other road agents. %
We evaluated our approach through a realistic simulator and tested it on a real SDV in both our private testing facility and on public roads, showing that \ours can be safely deployed in the real world. %

Our method also has some limitations.
For instance, the predicted distribution is not guaranteed to contain collision-free trajectories in all cases.
However, predicted trajectories improve by adding diverse training data, increasing the chance of having a collision-free trajectory (see Fig.~\ref{fig:ablation:ilk}).
Moreover, even when a collision is expected, our model can increase the time-to-collision, increasing the chances of finding a collision-free trajectory in subsequent re-planning cycles, before the potential collision occurs.
Also, our model does not guarantee scene consistency. Given the positive results in simulation and on the road, we speculate that the full scene consistency is not essential to improve the safety of our SDV planning approach.
Another limitation is that, even though our method does not rely on a rule-based trajectory generator, we use a handcrafted cost for the trajectory selection. 
Still, the cost is used only for trajectory selection rather than trajectory generation.
Finally, while we acknowledge that our closed-loop results are not reproducible due to the reliance on proprietary datasets and simulation tooling, we believe that our experiments clearly demonstrate the benefits of 
our approach,
not only in our closed-loop and on-road tests, but also via competitive prediction results on the Lyft Motion Prediction dataset.

\section*{Acknowledgments}
\noindent
We thank the whole Woven Planet, and in particular the research, motion planning, SimOps, TechOps and release teams, for support and discussions.
We also thank Nicolas Carion and Yann LeCun for discussions.

\bibliographystyle{abbrv}
{\small
\bibliography{main}  %
}

\clearpage
\appendix
\section{Appendix}\label{sec:appendix}

\subsection{On-road evaluation (attached video)}
\label{appendix:on_road}

In the supplementary video, available on the project webpage at \url{\shorturl}, we present the SDV driving \textbf{in full autonomy mode} in the streets of Palo Alto (CA), in the following scenarios:
\begin{itemize}[leftmargin=2em,topsep=0em,itemsep=0pt]
    \item \textbf{Yielding to cyclists at a stop intersection}: SDV stops at an unregulated stop intersection, waits for cyclists to pass and safely proceeds.
    \item \textbf{Left turn and passing a parked unloading truck}: SDV stops at a regulated intersection following a lead vehicle and then executes left turn while also nudging a parked unloading truck.
    \item \textbf{Cut-in maneuver}: A vehicle merges into SDV lane from an adjacent lane. SDV comfortably slows down.
    \item \textbf{Stop and left turn}: The SDV comes to a safe stop at the intersection before proceeding to make a left turn
    \item \textbf{Urban driving in dense traffic}: Multiple stretches of dense urban driving in full autonomy mode without any safety driver interventions.
\end{itemize}
The above on-road test was executed without using rule-based planners.
Note that the video shows agent trajectory predictions from a standalone integrated system for visualization purposes, which we plan to update with \ours predictions.
Our approach correctly handles challenging scenarios and safely drives on public roads.
This includes complex situations and intersections where many other road agents are present.
The road deployment confirmed that the planner was comfortable and felt safe for the SDV passengers.

\subsection{Implementation Details}\label{appendix:details}

\subsubsection*{Architectural details}\label{appendix:architecture}

As input to the model, we use $K_a = 1$ second of agent history, while we use $K_s = 3$ seconds of SDV history represented by moving/stationary information.
The model predicts $N = 10$ SDV trajectories with $T_s = 45$ steps
and $M = 5$ agent trajectories with $T_a = 30$ steps, both at 10 Hz.

We use 3 layers of PointNet for encoding map and road agents polylines with 128 dimensions.
The outputs are projected to $256d$ vectors and fed to the Transformer.
The learnable query embeddings for SDV and road agents have the same size.
We use $1024d$ feed-forward networks (FFN) inside the Transformer.
Both Transformer encoder and decoder have 3 layers.
The FFNs used as trajectory decoder are composed of a hidden $512d$ layer and an output layer with shape dependent of the prediction horizon.
In total the network has 5.7M parameters.

\subsubsection*{Loss weights}\label{appendix:loss}

Here we recap the matching cost and loss equations from the main draft.

We compute matching cost for each trajectory then select one trajectory according to
\begin{equation}
    i^* = \argmin_i \; \loss_\text{IL}^i + \lambda (1- p_i), \quad\;
    \text{where} \;\; \loss_\text{IL}^i = \sum_{t=1}^{T} ||\tau_t^i - \hat{\tau_t}||_1 + \beta \loss_\text{reg}^i,
    \label{eq:appendix:assignment_cost}
\end{equation}
where $p_i$ is the predicted probability for the trajectory $\tau^i$,
$\hat{\tau}$ is the ground truth trajectory,
$\lambda$ and $\beta$ are weighting factors,
and $\loss_\text{reg}$ is a regularization loss.
$\lambda$ controls the weight of the assignment probability compared to the imitation loss, while $\beta$ controls the smoothness of the predicted trajectories.
We empirically set
$\lambda_\text{SDV}=1$,
$\beta_\text{SDV}=1$,
$\lambda_\text{agent}=0.04$,
$\beta_\text{agent}=0$. 
We note that we do not apply regularization to the agent trajectories since they do not have to be executed. We also scale $\lambda_\text{agent}$ taking into account that the agent outputs are not regularized and passed through a kinematic decoder. 

We minimize the following loss
\begin{equation}
    \loss = \loss_\text{IL}^{i^*} + \mu \loss_{\text{NLL}}^{i^*}, \quad\; \text{where} \;\;  \loss_\text{NLL}^{i^*} = -\log \, p_{i^*},
    \label{eq:appendix:main_loss}
\end{equation}
that takes into account the selected trajectory $\tau^{i^*}$ and combines the imitation and the matching loss.
Similarly to DETR, we set $\mu = \lambda$ for both SDV and agents.

Overall, our total loss combines both SDV and agents losses, optimized simultaneously:
\begin{equation}
  \loss_\text{total} = \loss_\text{SDV} + \alpha \loss_\text{agent},
\end{equation}
where $\alpha$ is agent loss coefficient. It is set to 10 in all experiments.

For perturbations we use the same hyperparameters as SafetyNet~\cite{Vitelli2021SafetyNetSP}.

\end{document}